\documentclass[lettersize,journal]{IEEEtran}
\usepackage{booktabs}
\usepackage[switch]{lineno}
\usepackage[T1]{fontenc}

\usepackage{multirow}
\usepackage{graphicx}
\usepackage{pifont}
\usepackage[normalem]{ulem}
\usepackage{amsmath}
\usepackage{amssymb}
\usepackage{comment}
\usepackage{array}
\usepackage[compatibility=false]{caption}
\usepackage{subcaption}
\usepackage{threeparttable} 
\usepackage{url}
\usepackage{cite}
\usepackage{xcolor}

\usepackage{booktabs, multirow, siunitx}
\usepackage[hidelinks]{hyperref}
\usepackage{orcidlink}

\begin{document}

\bstctlcite{IEEEexample:BSTcontrol}

\title{AdROD: HyperNetwork-based Adversarially Robust Object Detection for Autonomous Driving}

\author{Yuting Wu\,\orcidlink{0009-0007-1510-6464}, Dongfang Guo\,\orcidlink{0000-0002-7464-0823}, Xiangzhong Luo\,\orcidlink{0000-0002-0758-2248}, Qun Song\,\orcidlink{0000-0002-3611-9404}, and Rui Tan\,\orcidlink{0000-0001-8441-9973}%
\thanks{This work was supported in part by the National Research Foundation, Singapore, and DSO National Laboratories under the AI Singapore Programme (AISG Award No. AISG4-GC-2023-006-1B); in part by the Ministry of Education, Singapore, under its Academic Research Fund Tier 2 (Award No. T2EP20225-0007); in part by City University of Hong Kong (Project No. 9610753); and in part by the JC STEM Lab of Future Energy Systems (Project No. 2025-0039). (\textit{Corresponding author: Rui Tan.})  }
\thanks{Yuting Wu, Dongfang Guo, Xiangzhong Luo, and Rui Tan were with Nanyang Technological University, Singapore. Qun Song was with the City University of Hong Kong, Hong Kong SAR, China. }
}

\markboth{IEEE Transactions on Computer-Aided Design of Integrated Circuits and Systems}%
{Wu \MakeLowercase{\textit{et al.}}: AdROD:HyperNetwork-based  Adversarially Robust Object Detection for Autonomous Driving}

\maketitle

\begin{abstract} 
Camera-based object detectors are vulnerable to physical adversarial attacks designed to suppress detections. While adversarial training and input purification offer some protection, they often overfit to specific attack distributions and fail on adaptive adversaries. This paper presents AdROD, an embedded, stochastic ensemble defense software designed for autonomous driving. AdROD employs {\em low-rank HyperNetworks}, which require only 1.6\% of the parameter footprint of standard HyperNetworks, to generate diverse detectors at a per-frame rate, making it impractical for attackers to obtain the deployed detectors in time. To further improve adversarial robustness, AdROD incorporates a novel \emph{functional diversity} mechanism, which couples stochastic weight updates with unique input-space transformations. We design two serving modes of AdROD that strike different trade-offs between robustness and runtime overhead: AdROD-I, a continuous protection mode for maximum resilience that leverages inter-detector disagreement to recover compromised detections, and AdROD-II, an on-demand mode triggered by kinematic discontinuities in object tracking. Through comprehensive evaluation with synthetic benchmarks, physically deployed adversarial patches, and end-to-end safety tests in the OpenCDA co-simulator, AdROD outperforms five baseline defenses and exhibits superior generalizability compared with the evaluated adversarial-training baselines, while maintaining real-time performance for safely stopping the vehicle at a stop sign instrumented with adversarial patches. 
\end{abstract}




\section{Introduction}
\label{sec:intro}

Camera-based object detectors built on deep neural networks (DNNs) are essential components of autonomous vehicle perception but are susceptible to physical adversarial attacks~\cite{RP_2_on_detector,FTE,wang2023does,SIB}. These attacks exploit model vulnerabilities and embed crafted perturbations into the scene to disturb predictions. In practice, adversaries often attach adversarial patches to critical road objects and suppress the victim vehicle's detection of these objects, posing severe safety risks~\cite{wang2023does, SIB, wang2024revisiting}. To address these threats, existing defenses include {\em adversarial training}, which augments training data with perturbed samples~\cite{zhang2019towards, amirkhani2023survey}, and {\em input purification}, which disrupts the perturbations through techniques such as compression and masking~\cite{LGS, liu2022SAC}. However, as these defenses are often designed for certain attack characteristics, such as those covered by the adversarial training samples, their robustness degrades on novel attacks or adaptive adversaries that actively optimize perturbations to bypass deployed defenses~\cite{DBLP:conf/nips/TramerCBM20}. 

To address this limitation, we propose AdROD (Adversarially Robust Object Detection), a stochastic defense that employs \emph{HyperNetworks}~\cite{ha2017hypernetworks} to generate an ensemble of diverse detectors from random noise at the per-frame rate. The predictions of the generated ensemble are then fused to produce robust detection results.
This design establishes a \emph{moving target defense}~\cite{jajodia2011moving, DBLP:conf/sensys/SongYT19}, in that the continuous variation of the detector ensemble renders the adversary's timely exploitation of the DNN gradients and decision boundaries impractical. Furthermore, AdROD is trained exclusively on benign data to remain agnostic to specific patch distributions, thereby reducing dependence on attack-specific training distributions.


However, standard HyperNetworks that generate the full weight space of the target DNN model scale poorly to support modern high-capacity object detectors. Specifically, the high dimension of the weight space leads to parameter explosion and destabilized training.
AdROD resolves this by introducing {\em low-rank HyperNetworks} that produce compact, low-rank updates to a pretrained base detector rather than full weight replacements~\cite{hu2021lora}. This design preserves knowledge contained in the factory model while drastically reducing memory overhead and promoting embedded deployments on resource-constrained platforms. Moreover, to mitigate attack effects that survive the {\em weight diversity} introduced by HyperNetworks, AdROD additionally enforces a novel \emph{functional diversity} by coupling each weight update with a unique, noise-driven input transformation. This dual-layer randomization forces the ensemble's detectors to react to out-of-distribution patterns differently, thereby reducing the transferability of adversarial patches.


With the ensemble generation capability as the basis, the next question is how to fuse the ensemble's detection results. Existing fusion approaches typically rely on rigid bounding-box score averaging or voting \cite{casado2020ensemble}, inherently trading benign accuracy for adversarial robustness. For instance, lowering the confidence thresholds may recover the objects missed due to attacks but inevitably increases the rate of benign false positives, as both attack-compromised true objects and spurious background predictions can receive low confidence scores. To navigate this trade-off, we design two serving modes. First, AdROD-I leverages inter-model disagreement to distinguish compromised detections from benign false positives. Second, to prioritize runtime efficiency, AdROD-II employs an on-demand serving strategy. Specifically, a low-overhead object tracker monitors the base detector under benign conditions and dynamically activates the ensemble defense upon abrupt object disappearances that signal potential attacks.


Our contributions are summarized as follows. {\bf First}, we propose AdROD, a stochastic ensemble defense driven by parameter-efficient \emph{low-rank HyperNetworks}. This design incurs only 1.6\% of the parameter footprint of standard HyperNetworks and stabilizes training.
{\bf Second}, by exploiting the computational headroom arising from the low-rank design, we introduce \emph{functional diversity}, a dual-layer randomization that couples weight diversity with input transformations for improved robustness. {\bf Third}, we introduce two specialized serving modes: a continuous protection mode (AdROD-I) optimized for maximum robustness, and an adaptive on-demand mode (AdROD-II) tailored for runtime efficiency. {\bf Fourth}, we conduct extensive evaluation from perception performance to end-to-end safety via the OpenCDA pipeline~\cite{xu2021opencda}.\footnote{Demo videos are available at \url{https://github.com/yutingwu-ntuiot/AdROD}.} AdROD consistently outperforms five representative baselines~\cite{guo2017countering, liu2022SAC, Objseeker,  tarchoun2023jedi, LGS} and exhibits superior generalizability over adversarial training, while its adaptive on-demand mode preserves the real-time throughput required for safe vehicle control in front of traffic signs instrumented with adversarial patches.

{\em Paper organization:} 
\textsection\ref{Background} introduces preliminaries and reviews related work.  
\textsection\ref{overview} states the research problem. 
\textsection\ref{approach} presents the design and training of AdROD. 
\textsection\ref{U2E} describes the two serving modes. 
\textsection\ref{evaluation} presents implementation details and evaluation results.
\textsection\ref{discussion} discusses the limitations and future work.
\textsection\ref{dis_and_conclude} concludes this paper.



\section{Preliminaries and Related Work}
\label{Background}

\subsection{Camera-based Object Detection and Adversarial Attacks}  
\label{detection_loss}  

Camera-based object detection forms the perceptual foundation for autonomous vehicles and subsequent navigation tasks. Modern detectors are broadly categorized into one-stage architectures (e.g., YOLO~\cite{jocher2020yolov5}), which perform localization and classification simultaneously, and two-stage architectures (e.g., Faster R-CNN~\cite{faster_rcnn}), which rely on region proposal networks. Both approaches ultimately utilize Non-Maximum Suppression (NMS) to filter redundant bounding boxes. Specifically, NMS suppresses all candidate boxes with a confidence score below a threshold $\tau$, and then iteratively removes overlapping boxes that exceed a predefined Intersection over Union (IoU) threshold $\gamma$, ensuring only the most prominent detection remains. We adopt a confidence threshold $\tau\!=\!0.3$ and an IoU threshold $\gamma\!=\!0.4$, which follow common settings in the literature (e.g., \cite{liu2022SAC}). While this paper primarily adopts YOLO as the design basis for AdROD due to its high efficiency, AdROD can also adopt Faster R-CNN as evaluated in \textsection\ref{evaluation}.

Despite DNN-based object detectors' state-of-the-art accuracy under benign conditions, they are vulnerable to physical adversarial attacks, including regional solid patches or stickers affixed to objects~\cite{RP_2_on_detector,DBLP:conf/ndss/He0ZZHB0024} and global semi-transparent overlays projected or placed on the objects~\cite{chen2019shapeshifter, guo2024invisible, zolfi2020translucent, DBLP:conf/ndss/SatoBC0CR24}. Among these, {\em adversarial patches} have received wide attention due to their simplicity and low cost~\cite{wang2024revisiting}.
Formally, an adversarial patch $\mathbf{P}$ modifies a benign image $\mathcal{X}$ into an adversarial image $\mathcal{X}_\text{adv}$ via binary mask $\mathbf{M}$ and spatial transformation $ \mathcal{A}$, yielding $\mathcal{X}_{\text{adv}} = \mathbf{M} \odot \mathcal{A}(\mathbf{P})+(1-\mathbf{M}) \odot \mathcal{X}$.
The attack design typically adopts the loss function \( \mathcal{L}_\text{hide} = \alpha_\mathrm{conf} \cdot \mathcal{L}_\text{conf} + \mathcal{L}_\text{colors} \), where $\mathcal{L}_\text{conf}$ measures the reduction in detection confidence for the target object (e.g., objectness loss), $\mathcal{L}_\text{colors}$ enforces constraints on the patch's visual realism and printability, and $\alpha_\mathrm{conf}$ is a weight. To improve attack robustness under varying environmental conditions, adversaries can use  Expectation Over Transformation (EOT)~\cite{athalye2018synthesizing}, which optimizes the patch under randomized transformations to mimic real-world shifts in object scale, spatial placement, illumination, and background context. For instance, the $\text{RP}_2$ attack \cite{RP_2_on_detector} adopts a uniform distribution of sign size;
the SysAdv attack \cite{wang2023does} adopts a distribution derived from real traces.

\subsection{Adversarial Defenses}
\label{sec:adversarial defense}
Existing defense approaches against adversarial attacks broadly fall into the following categories. A common solution is \textit{adversarial training}~\cite{zhang2019towards,im2022adversarial}. While adversarial training introduces no additional inference latency, this approach can overfit to attack-specific training distributions and exhibits limited generalization to novel threats. Furthermore, exhaustive anticipation of all potential attacks is not practical. \textit{Input purification and patch removal} techniques preprocess the input to disrupt or remove adversarial perturbations. For example, Local Gradient Smoothing (LGS)~\cite{LGS} filters high-gradient regions, Jedi~\cite{tarchoun2023jedi} localizes adversarial patches using entropy-based cues, and Segment-And-Complete (SAC)~\cite{liu2022SAC} segments and completes suspicious patch regions. Other recent works follow a similar localization-and-removal principle. Adversarial Pixel Masking~\cite{DBLP:conf/mm/ChiangCW21} learns a MaskNet to mask out patch-like perturbations for pre-trained object detectors. PatchZero~\cite{DBLP:conf/wacv/XuXZCN23} detects adversarial pixels and zeros out the detected patch region by repainting it with mean pixel values. Attention-based defenses~\cite{DBLP:conf/iccps/RossoliniBB24} use internal attention signals to track and mask malicious regions in multi-frame vision applications. These methods are effective when the adversarial regions can be reliably identified. However, they still rely on patch localization or masking rules, which may be challenged by adaptive attacks, low-frequency or smooth patches, irregular patch shapes, and patches that overlap heavily with the target object. For instance, as shown in \textsection\ref{evaluation}, customized smooth textures can bypass LGS's gradient-based filtering; visually realistic patches with irregular shapes can defeat SAC, which is primarily designed to address square patches with high-frequency textures. Overall, these defenses either depend on attack-specific training data or apply fixed preprocessing rules, making them difficult to generalize across diverse and adaptive physical attacks.


Beyond attack-specific or static defenses, another category of research improves robustness through model diversity or runtime randomness. \textit{Ensemble-based defenses} use multiple models during training or inference and have shown effectiveness mainly in image classification settings~\cite{DBLP:conf/iclr/TramerKPGBM18, DBLP:conf/icml/PangXDCZ19, DBLP:conf/nips/YangZDIGTWB020}. However, many of them rely on adversarial training or fixed model sets, which may limit their robustness against white-box adaptive attackers. Directly applying such ensembles to complex object detection also raises practical issues, including high inference latency and difficulty in balancing robustness with benign accuracy, as discussed in \textsection\ref{c1}. \textit{Stochastic defenses} introduce randomness to disrupt attack transferability, such as using dropout~\cite{wang2018defensive} or randomly pruning activations~\cite{dhillon2018stochastic} during inference. HyperNetwork-based dynamic ensembles~\cite{song2022sardino, xu2025dynamic} combine \textit{ensemble diversity with runtime stochasticity} by generating different model instances from random noise and are therefore the prior designs closest to AdROD. However, directly applying this approach to camera-based object detection remains challenging, as discussed next.

\subsection{HyperNetworks and Their Limitations}

Standard HyperNetworks~\cite{ha2017hypernetworks} provide a fast way for dynamic parameter generation. They construct the weights of a target DNN model from random noise. Formally, a generator \( \mathcal{G} \) is assigned to each layer of the target model, and maps a shared random vector \( \mathbf{z} \in \mathbb{R}^{1 \times d} \) to a new weight matrix \( \mathbf{\Theta}' \), which then replaces the original weights \( \mathbf{\Theta} \) of that layer. 


Song et al.~\cite{song2022sardino} utilized HyperNetworks to protect single-label image classifiers, while Xu et al.~\cite{xu2025dynamic} extended this scheme to LiDAR-based object detection. However, directly adapting HyperNetworks to camera-based object detection faces two challenges. First, LiDAR attacks in~\cite{xu2025dynamic} typically inject sparse, spurious points while largely preserving the victim vehicle's geometric structure. In contrast, physical attacks on camera-based object detection use intrusive patches that corrupt target-related visual features. Consequently, weight diversity alone in~\cite{xu2025dynamic} is insufficient to deal with physical attacks on camera-based object detection, as shown in \textsection\ref{evaluation}.  Second, scaling standard HyperNetworks to modern convolutional object detectors causes substantial parameter growth and overhead increase, as quantified in \textsection\ref{overview}. The resulting larger optimization space also makes it harder to jointly balance ensemble diversity and benign detection accuracy.

\section{Problem Statement} 
\label{overview}

\subsection{Threat Model}
\label{sec:define}

\textbf{Attack objectives and scope.} We consider an autonomous driving scenario in which a vehicle may encounter physical adversarial attacks. An attack is deemed successful if it subverts detection in any of the following ways: (1) the target object is not detected, meaning no predicted bounding box overlaps with the ground truth; (2) the object is misclassified; or (3) the object is poorly localized (i.e., low IoU). These criteria collectively reflect detection quality across object presence, classification, and localization. Accordingly, this work focuses on attacks that compromise the detection of existing objects.

\textbf{Attacker capabilities.} Following prior studies~\cite{RP_2_on_detector, SIB, wang2023does}, we consider two attacker knowledge settings. Under a weak adversarial setting, the attack is optimized against an undefended vanilla model. Under a strong adversarial setting, the attacker possesses full white-box knowledge of the AdROD pipeline to optimize the adversarial pattern, with the exception of the runtime stochastic noise vectors. This is consistent with Shannon's maxim~\cite{shannon1949communication}, under which the adversary is assumed to know the system except for private runtime randomness.


\textbf{Attack implementation.}
We consider physically plausible adversarial patterns that are visible to the victim camera. These include patterns attached to target objects (e.g., stop signs, persons, and vehicles), projected onto targets (e.g., stop signs), or placed near the target (e.g., traffic light). We assume that the attacker cannot modify the digital image stream or arbitrarily perturb every pixel.

\begin{figure}[t]
  \centering
  \includegraphics[width=1.0\linewidth]{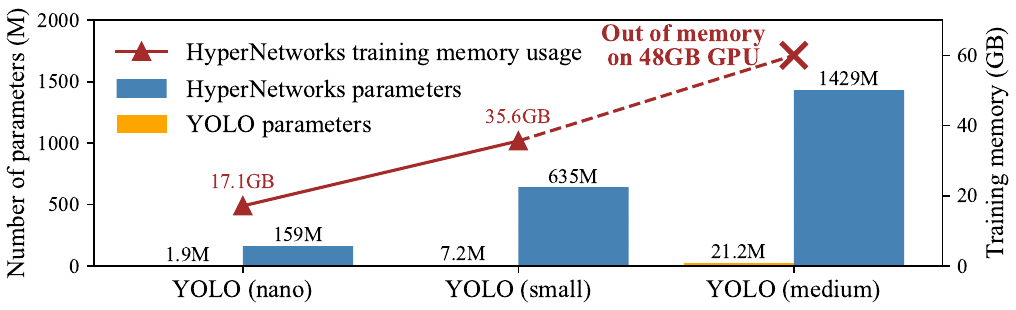}
  \vspace{-1.5em}
  \caption{Large parameter spaces and training memory requirements of HyperNetworks for three YOLO versions.}
  \label{fig:ob_1}
\end{figure}

\subsection{Challenges of HyperNetwork-based Defense}
\label{c1} 


We aim to design a HyperNetwork-based defense that generates an ensemble of diverse detectors at runtime and fuses their outputs to produce a reliable final result. If the attack designed against certain detectors transfers poorly to the generated detectors, robust object detection can be achieved. However, the defense design faces the following system challenges.


\begin{figure}
  \includegraphics[width=1\linewidth]{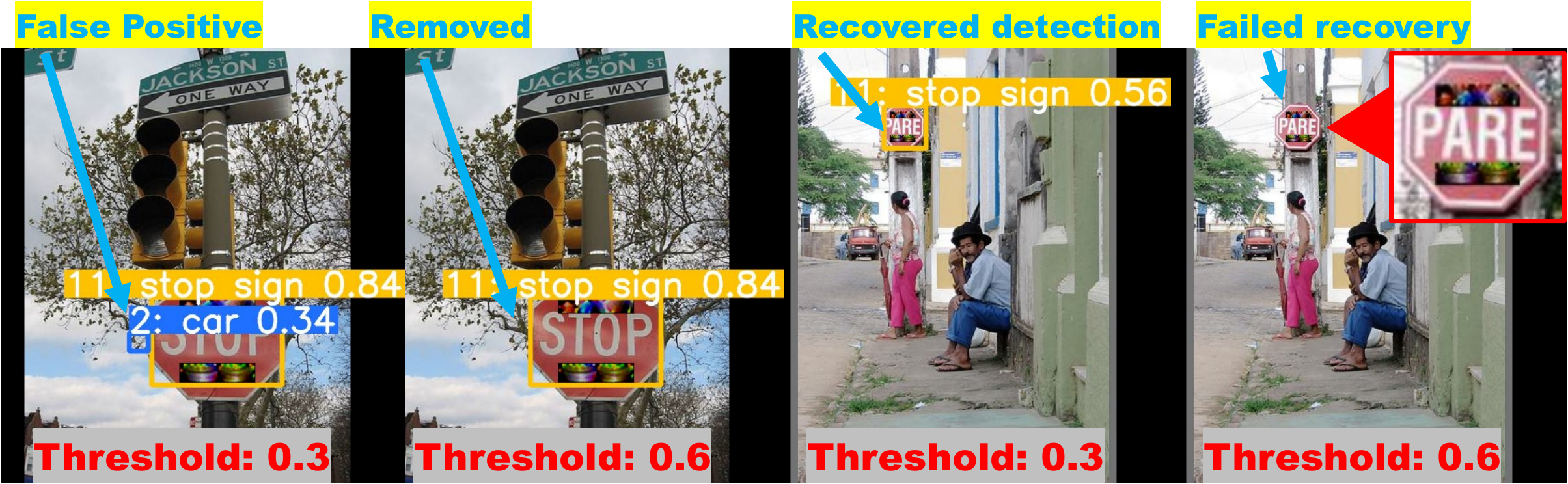}
  \caption{Illustration of the trade-off between robustness and FPR under an adversarial attack on a stop sign. Raising the confidence threshold removes an undesired false positive (left) but hinders the recovery of the victim stop sign that would otherwise be detected at the lower threshold (right). Bounding boxes for other correct detections are omitted for readability.}
    \label{fig:ob_3}
  \vspace{-1em}
\end{figure}

\textbf{HyperNetworks scalability.} Standard HyperNetworks exhibit a significant parameter overhead relative to the target model. As illustrated in Fig.~\ref{fig:ob_1}, standard HyperNetworks for target models of YOLO-nano, YOLO-small, and YOLO-medium require 67$\times$, 88$\times$, and 84$\times$ more parameters than their respective target models. This leads to memory problems and poor training convergence.
A new design for parameter-efficient HyperNetworks is required.

\textbf{Robustness vs. false positive rate (FPR).}
An inherent trade-off exists between adversarial robustness and the benign FPR. Both victim objects and benign false positives (FPs) often exhibit low detection confidence. As illustrated in Fig.~\ref{fig:ob_3}, naive mitigation strategies, such as lowering confirmation thresholds to recover a compromised stop sign, inevitably increase the retention of undesired FPs. Improving the trade-off requires a novel, discriminative metric to distinguish attack-related objects from benign FPs. This paper identifies {\em objectness variance} (cf.~\textsection\ref{sec:rod}) as such a novel metric.

\textbf{Serving efficiency.} A continuous ensemble defense may incur high overhead. For instance, YOLO-small on a Jetson AGX Xavier achieves a throughput of 66.9 frames per second (FPS) at $640\times640$ resolution, whereas executing a continuous ensemble of 10 generated models drops to 5.4 FPS. Trading computational resources for safety is necessary, but unnecessary overhead should be avoided when the vehicle operation is safe. To this end, we seek a proactive yet low-overhead mechanism for detecting attack onset and activating ensemble defense on demand (cf.~\textsection\ref{m2}).

\textsection\ref{approach} and \textsection\ref{U2E} address these challenges through low-rank generation, functional diversity, and two deployment modes.


\section{A New HyperNetwork Design for AdROD}
\label{approach}

 \begin{figure}
    \centering
    \includegraphics[width=1\linewidth]{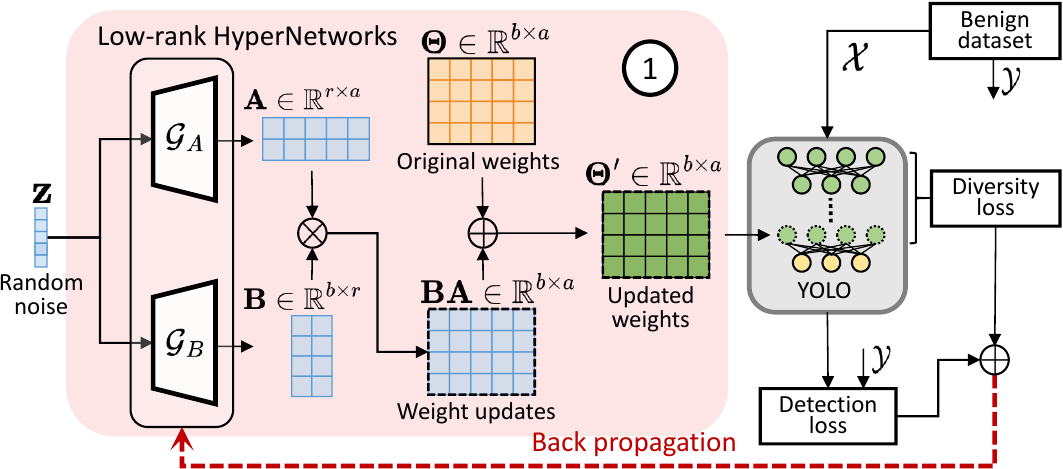}
    \caption{AdROD training overview. The green circles in the YOLO architecture indicate weights updated by the low-rank HyperNetworks ($\mathbf{BA}$), while the yellow circles represent the original weights ($\boldsymbol{\Theta}$) retained from the base detector.}
    \label{fig:overview}
  \vspace{-1em}
\end{figure}

AdROD resolves scalability bottlenecks and reduces attack transferability via low-rank HyperNetworks (Fig.~\ref{fig:overview}-\ding{172}, \textsection\ref{sec:low-rank}) and functional diversity (Fig.~\ref{fig:overview}-\ding{173}, \textsection\ref{sec:diverse}), respectively. Unlike adversarial training, the proposed architecture is optimized exclusively on benign images $\mathcal{X}$ and their object annotations $\mathcal{Y}$, driven jointly by an average detection loss and a weight diversity loss across the generated ensemble.

To illustrate the contribution of individual architectural components, we use the EOT-optimized SysAdv attack~\cite{wang2023does} on COCO~\cite{lin2014microsoft} stop sign instances as a running example.
Unless stated otherwise, ten generated models are fused {\em affirmatively} \cite{casado2020ensemble} (i.e., retaining any detection from any model to maximize robustness) and followed by NMS. We evaluate robustness via {\em attack success rate} (ASR), defined as the percentage of frames where the attack succeeds, and {\em benign detection accuracy} measured by mean average precision (mAP) on the attack-free COCO validation dataset. Paragraphs with a heading mark of ``\textreferencemark'' report results of the running example.

\subsection{Low-rank HyperNetworks}
\label{sec:low-rank}

\begin{figure}
    \centering
    \includegraphics[width=1.0\linewidth]{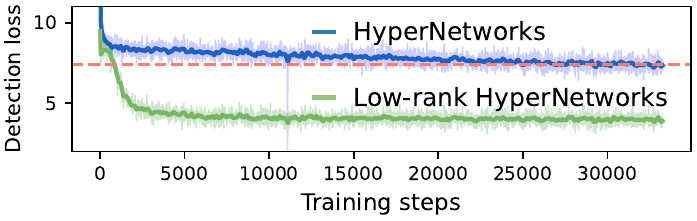}
    \vspace{-1.5em}
    \caption{Low-rank HyperNetworks achieve lower training detection loss compared with standard HyperNetworks.}
    \label{fig:4_1}
    \vspace{-1em}
\end{figure}

AdROD integrates Low-Rank Adaptation (LoRA)~\cite{hu2021lora} with HyperNetworks for scalability. While recent works have explored LoRA-HyperNetworks integrations for cross-task generalization in large language models and physics-informed networks~\cite{lv2024hyperlora, majumdar2023hyperlora}, AdROD specifically deploys this architecture for adversarial robustness in object detection. Instead of generating the weights of a target model directly, the proposed low-rank HyperNetworks generate updates to the weights of a factory-designed {\em base detector} that offers validated benign accuracy. Specifically, for a base detector's layer with pre-trained weights $\boldsymbol{\Theta} \in \mathbb{R}^{b \times a}$, the system employs two smaller generators, $\mathcal{G}_A$ and $\mathcal{G}_B$, to produce low-rank matrices $\mathbf{A} \in \mathbb{R}^{r \times a}$ and $\mathbf{B} \in \mathbb{R}^{b \times r}$, where the rank $r \ll \min\{a, b\}$. Their product, \(\mathbf{B} \mathbf{A} \in \mathbb{R}^{b \times a}\), updates the original weights as $ \boldsymbol{\Theta}' = \boldsymbol{\Theta} + \mathbf{BA}$. This decomposition preserves the learned knowledge in $\boldsymbol{\Theta}$ while reducing the trainable parameter footprint by up to two orders of magnitude. The resulting training efficiency mitigates the optimization instability observed with standard HyperNetworks. In our implementation, we use the same rank $r$ for all LoRA-updated layers to keep the design simple.


\textreferencemark~\textbf{Impact of rank $r$.}
From empirical evaluation across rank settings of $\{1, 4, 8, 16, 32, 64\}$, we observe that $r\!=\!1$ severely degrades mAP, whereas $r\!=\!64$ causes training instability. Although mAP peaks at $r\!=\!32$, increasing $r$ degrades adversarial robustness. To achieve a satisfactory balance between accuracy and robustness, we adopt $r\!=\!4$. With this setting, the low-rank HyperNetworks for the YOLO-small base detector require only 1.6\% of the parameters of standard HyperNetworks, reducing the parameter count from 635 million to about 10 million. They also achieve 45.9\% lower detection loss than standard HyperNetworks, as shown in Fig.~\ref{fig:4_1}.

\subsection{Diversifying Models for Robustness}
\label{sec:diverse}
To comprehensively disrupt attack transferability, AdROD establishes \textit{functional diversity} by integrating \textit{weight diversity} with input transformations. Let \( K \) denote the number of models in the ensemble (i.e., ensemble size), and let $\boldsymbol{\Theta}_{k,n}'$ denote the weight matrix of the $n$-th updated layer of the $k$-th model, where $1 {\le} k {\le} K$ and $1 {\le} n {\le} N$. We flatten $\boldsymbol{\Theta}_{k,n}'$ into a vector $\boldsymbol{\theta}_{k,n}'\in \mathbb{R}^{1 \times ab}$. The weight diversity score $ \gamma_n $ across $K$ models is defined via Pearson correlation:
\begin{equation*}
    \gamma_n = \frac{1}{\binom{K}{2}} \sum_{1 \leq k_1 < k_2 \leq K} \left| \text{Pearson correlation}(\boldsymbol{\theta}_{k_1,n}', \boldsymbol{\theta}'_{k_2,n}) \right|.
\end{equation*}
The low-rank HyperNetworks are jointly optimized using the loss $\mathcal{L} = \frac{1}{K} \sum_{k=1}^{K} \mathcal{L}_{\text{det},k} + \frac{\beta}{N} \sum_{n=1}^N \gamma_n^2$, where $\mathcal{L}_{\text{det},k}$ is the primary detection loss and $\beta \ge 0$ dictates the diversity penalty. Note that the training ensemble used to compute the diversity loss is kept small to reduce GPU memory usage during joint optimization, which does not need to be matched with the serving ensemble size tuned for robust detection.

\begin{figure}
    \centering
    \begin{subfigure}[b]{0.49\linewidth}
      \includegraphics[width=\linewidth]{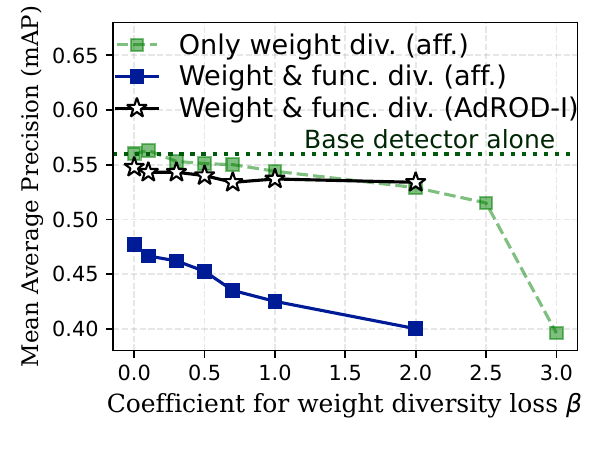}
        \vspace{-2em}
        \caption{Impact of $\beta$ on mAP}
        \label{fig:beta_on_map}
    \end{subfigure}
    \begin{subfigure}[b]{0.49\linewidth}
      \includegraphics[width=\linewidth]{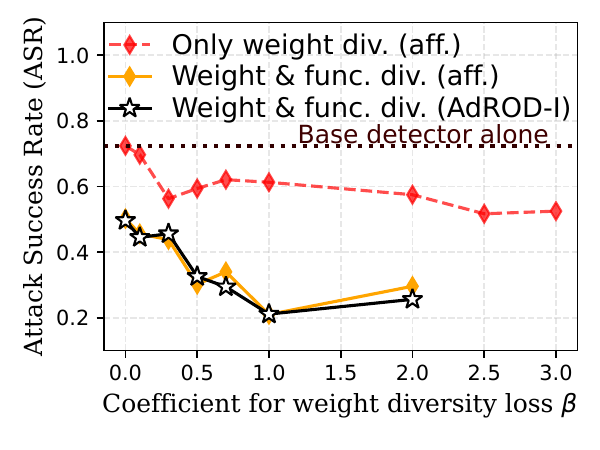}
        \vspace{-2em}
        \caption{Impact of $\beta$ on ASR}
        \label{fig:beta_on_asr}
      \end{subfigure}
      \vspace{-1em}
    \caption{Comparison of ensemble detection performance under affirmative (aff.) fusion and the fusion of AdROD-I. Incorporating functional diversity (func. div.) improves robustness compared with using weight diversity only, while AdROD-I further mitigates benign detection accuracy degradation.} 
    \label{fig:ap_3}
        \vspace{-1em}
\end{figure}


\textreferencemark~\textbf{Impact of coefficient $\beta$.}  Fig.~\ref{fig:beta_on_asr} shows that increasing $\beta$ initially lowers ASR but saturates for $\beta \ge 2.5$. Meanwhile, Fig.~\ref{fig:beta_on_map} shows that the mAP drops sharply. Weight diversity alone is therefore insufficient, motivating an orthogonal, input-level countermeasure.


\subsubsection{Functional Diversity}
\label{subsubsec:func-diversity}


To further reduce attack transferability, AdROD introduces functional diversity by using the HyperNetwork's random noise vector $\mathbf{z} \in \mathbb{R}^{1 \times d}$ to jointly generate the detector's weight updates and parameterize an input-space transformation $\mathbf{T}$ during both training and serving, i.e., $\mathcal{X}' = \mathbf{T}(\mathcal{X} \! \mid \! \mathbf{z})$. Thus, each generated detector acts as a ``decoder'' for its own ``encoded'' transformed input, reducing attack transferability across ensemble members.



\begin{figure}
    \centering
    \includegraphics[width=\linewidth]{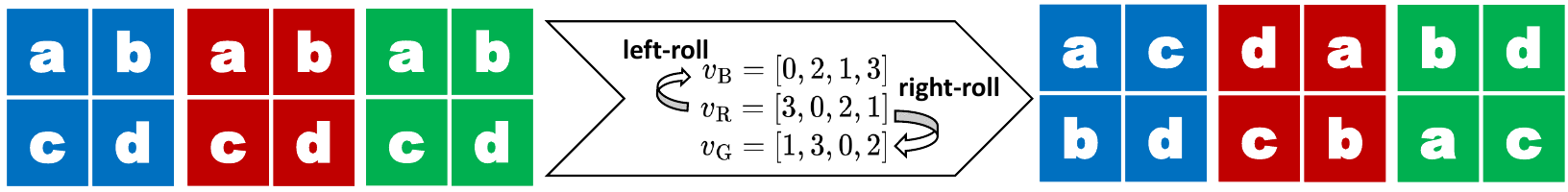}
    \includegraphics[width=\linewidth]{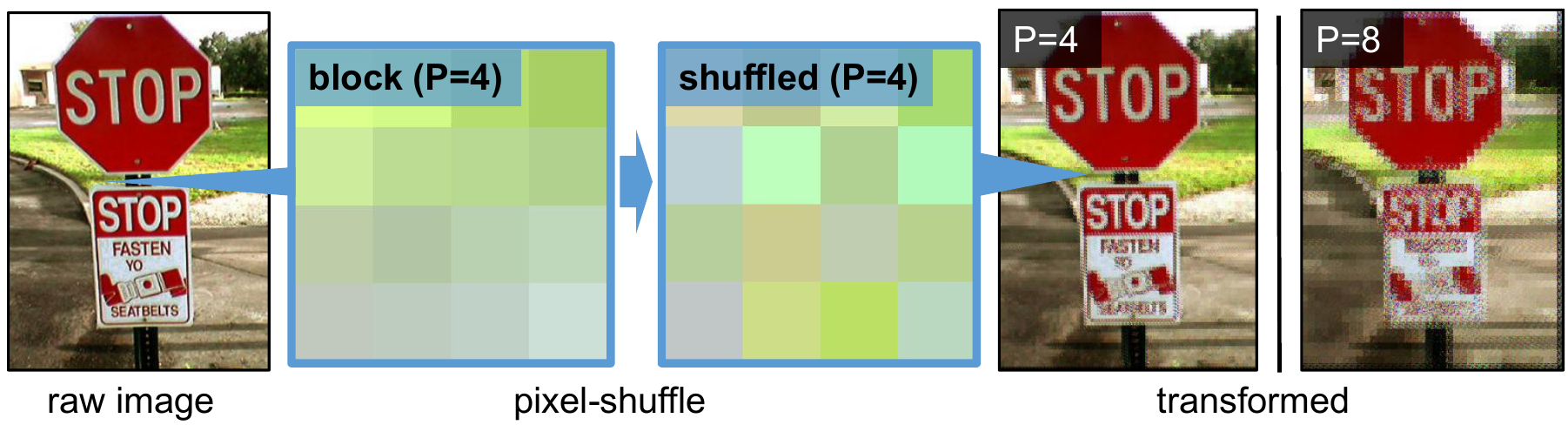}
    \caption{Top: Permutation orders and permutation process for $P$=2. Bottom: Effects of pixel shuffling for $P$=4 and $P$=8.}
    \label{fig:Pixel_shuffle}
        \vspace{-1em}
\end{figure}

Rather than relying on computationally heavy cryptographic operations, we adopt block-wise pixel shuffling \cite{aprilpyone2021block, maungmaung2023efficient} to attain high-throughput spatial distortion. Specifically, the pipeline divides the input image $\mathcal{X}$ into several non-overlapping $P{\times}P$ blocks and shuffles each block following a {\em permutation order}. The permutation order of the red channel, \( \mathbf{v}_R {=} (v_1, \ldots, v_{P^2}) \), is derived by average-pooling $\mathbf{z}$ with a kernel of \( \frac{d}{P^2} \), where $P^2 ~{<}~d$, then ranking the pooled vector. To decorrelate the color channels, the permutation orders for green and blue channels, $\mathbf{v}_G$ and $\mathbf{v}_B$, are generated by right- and left-rolling $\mathbf{v}_R$, respectively.  Fig.~\ref{fig:Pixel_shuffle} illustrates this process.

\textreferencemark~\textbf{Effectiveness of functional diversity.} Evaluating the spatial block size $P \in \{1, 2, 4, 5, 8\}$ reveals a trade-off between patch distortion and feature preservation. Larger $P$ values induce image distortion, which more strongly distorts the adversarial patch but degrades benign detection accuracy. With $P\!=\!4$, which balances robustness and benign accuracy, functional diversity brings an average reduction of 26.3 percentage points in ASR for $\beta \in [0, 2.0]$ compared with weight diversity alone, as shown in Fig.~\ref{fig:beta_on_asr}. However, as shown in Fig.~\ref{fig:beta_on_map}, this robustness enhancement incurs an average mAP drop of 10.5 percentage points. This drop stems from the affirmative fusion rule, which accumulates FPs across the diverse ensemble, compounded by transformation-induced false negatives (FNs). In the next section, we devise an attack-aware fusion scheduler to recover benign detection accuracy without compromising the attained robustness.

\section{Two Serving Modes of AdROD}
\label{U2E}

This section presents two serving modes: AdROD-I employs ensemble disagreement as an attack indicator to strike a better balance between adversarial robustness and benign detection accuracy; AdROD-II exploits the kinematic discontinuity of attack effectiveness observed during the victim vehicle’s movement to reduce unnecessary ensemble execution.

\subsection{AdROD-I}
\label{sec:rod}

\begin{figure}
    \centering
    \includegraphics[width=1\linewidth]{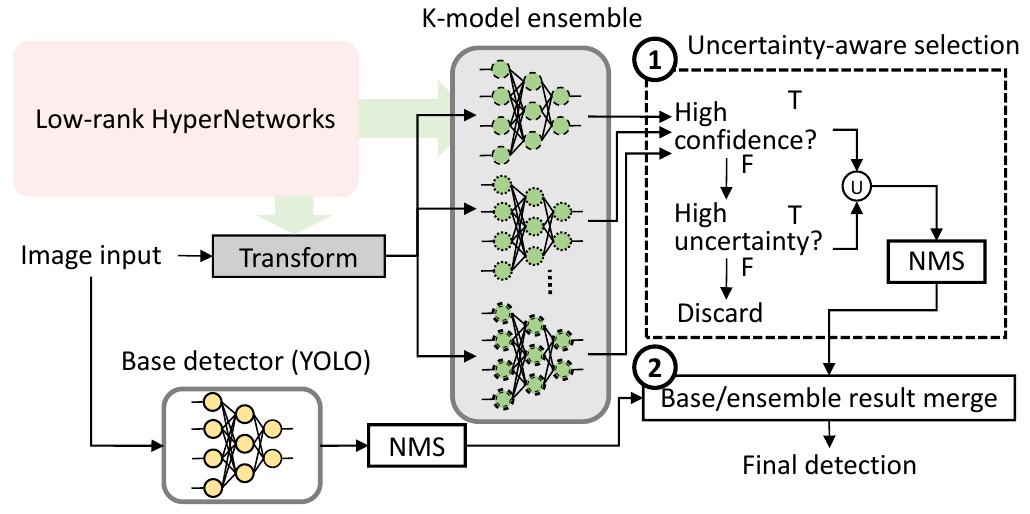}
    \vspace{-1.5em}
    \caption{AdROD-I workflow.}
    \label{fig:rod_overview}
       \vspace{-1em}
\end{figure}

AdROD-I is built on two ideas to address the increased benign FPs and FNs caused by the ensemble defense. First, it uses an uncertainty indicator to distinguish benign FPs from predictions associated with victim objects (see Fig.~\ref{fig:rod_overview}~\ding{172}). Second, it merges the ensemble results with the base detector results to correct FNs caused by the image distortion introduced by functional diversity (see Fig.~\ref{fig:rod_overview}~\ding{173}). We continue to use the running example to illustrate the first idea.


To formalize the attack indicator, we first elaborate on some internal details of the object detector. Given an input image, the detector produces predictions for densely predefined spatial anchors across the image, each representing a candidate object hypothesis at a particular spatial location and scale. For each spatial anchor, the detector generates a prediction vector $\mathbf{b} = [x, y, w, h, o, \mathbf{p}]$, where $(x, y, w, h)$ encodes the bounding-box geometry, $o$ represents the scalar \textit{objectness} score, and $\mathbf{p} = [p_1,\ldots,p_C] \in [0,1]^C$ contains the per-class confidence scores over $C$ detection classes. The final detection confidence is computed as $o \cdot \max_c p_c$. We hypothesize that, when adversarial transferability is reduced, physical adversarial patches cannot suppress all generated detectors consistently. As a result, predictions around an attacked genuine object exhibit higher disagreement across the ensemble, whereas benign FPs tend to appear more consistently. To quantify this difference, AdROD-I computes the variance of $o$ across the $K$ ensemble members for each spatially aligned anchor. For each image frame, it then sorts the aligned anchor predictions in descending order of \textit{objectness variance}  to obtain a localized \textit{uncertainty rank}, where rank 1 indicates the highest disagreement. 

\begin{figure}
    \centering
    \begin{subfigure}[b]{0.565\linewidth}
        \includegraphics[width=1.0\linewidth]{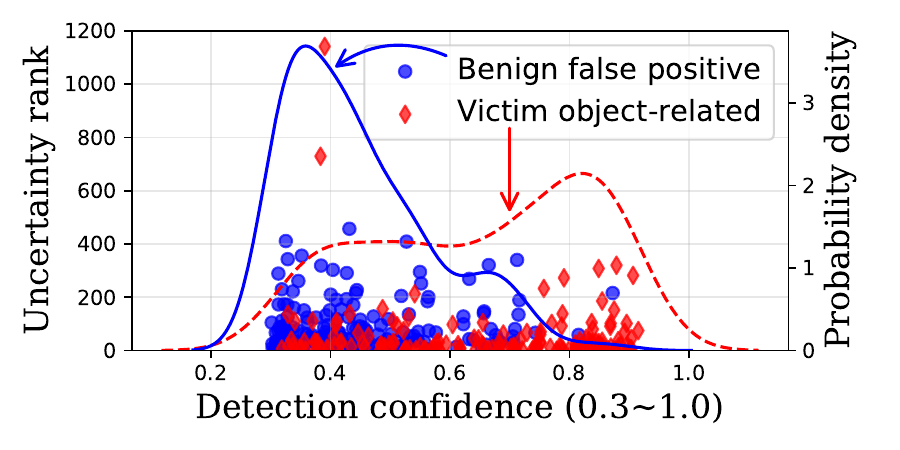}
        \vspace{-1.75em}
        \caption{Scatter plot of uncertainty rank vs. detection confidence and their probability density functions (PDFs)}
        \label{fig:pdf}
    \end{subfigure}
    \hfill
    \begin{subfigure}[b]{0.38\linewidth}
        \includegraphics[width=1.0\linewidth]{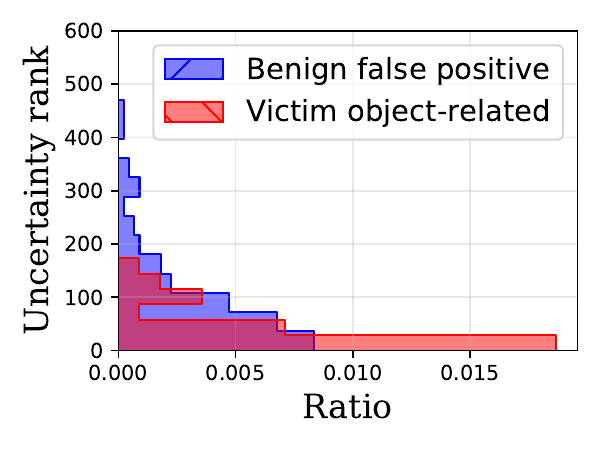}
         \vspace{-1.75em}
        \caption{Histogram of uncertainty rank (detection confidence $\in$ [0.3,0.5])}
        \label{fig:hist}
    \end{subfigure}
    \vspace{-0.5em}
    \caption{Distributions of uncertainty rank and detection confidence for benign FPs and victim objects.}
    \label{fig:approach_3}
        \vspace{-1em}
\end{figure}

\textreferencemark~Fig.~\ref{fig:pdf} shows the scatter plot of uncertainty rank versus detection confidence, where each point corresponds to a detection. This scatter plot merges results from all tested frames and distinguishes those corresponding to victim objects and benign FPs. It also shows the probability density functions (PDFs) of detection confidence for two cases. From the plot, high-confidence detections are rarely FPs, while low-confidence ones may be either. From Fig.~\ref{fig:hist}, for detections with confidence below 0.5, victim object-related detections are concentrated at low uncertainty ranks, showing that the attack likely increases objectness variance and disagreements among ensemble members.
These observations motivate our design of AdROD-I as follows.


\textbf{\ding{172} Uncertainty-aware selection.} First, from all anchors' box predictions $\mathcal{B}$ given by all detection models in the ensemble, we select the high-confidence predictions (denoted by $\mathcal{B}_\text{confident}$), which are likely benign and rarely false positives (cf. Fig.~\ref{fig:pdf}), and the low-confidence but top-$Q$ most uncertain predictions (denoted by $\mathcal{B}_\text{victim}$; cf. Fig.~\ref{fig:hist}), where $Q$ is the uncertainty quota controlling the number of retained victim object candidates:
\begin{align*}
\mathcal{B}_{\text{confident}} &= \{ \mathbf{b}\in\mathcal{B} \mid \operatorname{conf}(\mathbf{b}) \ge \tau_c \}, \\ \mathcal{B}_{\text{victim}} &= \{ \mathbf{b}\in\mathcal{B} \mid \operatorname{conf}(\mathbf{b}) < \tau_c \land \operatorname{rank}_u(\mathbf{b}) \le Q \}, 
\end{align*}
where $\operatorname{conf}(\mathbf{b})$ extracts $\mathbf{b}$'s detection confidence, and $\operatorname{rank}_u(\mathbf{b})$ denotes the uncertainty rank. We combine the above two sets and apply NMS to form the ensemble's detection result $\mathcal{D}_\text{ens}$, i.e., $\mathcal{D}_{\mathrm{ens}} = \operatorname{NMS} \left( \mathcal{B}_\text{confident} \cup \mathcal{B}_\text{victim} \right)$.
The NMS adopts its default confidence threshold $\tau$. We set $\tau_c > \tau$, so that we impose a stricter requirement to confirm an anchor-box prediction as benign for joining $\mathcal{B}_\text{confident}$. This helps reduce benign FPs. A greater $Q$ enhances victim recovery but also increases the risk of including benign FPs within $\mathcal{B}_\text{victim}$.


\textbf{\ding{173} Base/ensemble result merge.} 
To resolve FNs caused by transformation distortion and the confidence threshold $\tau_c$, we merge $\mathcal{D}_{\text{ens}}$ with the base detector's results on untransformed inputs $\mathcal{D}_{\text{base}}$. If two detections  \(\mathbf{d}_{\text{base}} \in \mathcal{D}_{\text{base}}\) and \(\mathbf{d}_{\text{ens}} \in \mathcal{D}_{\text{ens}}\) lack \textit{sufficient spatial overlap} (i.e., $\text{IoU} \!< \!0.6$), both are retained to recover benign objects detected only by the base detector and victim objects recovered by the ensemble. For pairs with \textit{substantial overlap}, the system applies the following resolution logic: (1) If predicted classes match, it retains only $\mathbf{d}_{\text{base}}$, prioritizing the base detector's superior bounding-box accuracy on untransformed inputs; (2) If classes diverge and $\mathbf{d}_{\text{ens}}$ possesses higher confidence, it retains $\mathbf{d}_{\text{ens}}$, treating it as a recovered victim object; (3) Otherwise, it conservatively retains both bounding boxes to prevent accidental suppression of ambiguous cases. 

\textreferencemark~\textbf{Effectiveness.} We set $\tau_c=0.5$ based on the results in Fig.~\ref{fig:approach_3}, and $Q=10$ based on the sensitivity analysis in \textsection\ref{sec:sensitive}. By sequentially executing steps \ding{172} and \ding{173}, as shown in Fig.~\ref{fig:ap_3}, AdROD-I reclaims the benign detection accuracy without compromising established adversarial robustness.

\subsection{AdROD-II}
\label{m2}

\begin{figure}
    \centering
    \includegraphics[width=1\linewidth]{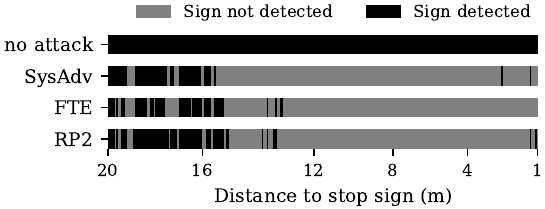}
    \vspace{-1.5em}
    \caption{Result trace of undefended YOLO when the vehicle approaches a stop sign with an adversarial patch.  The patch is digitally applied to maximize attack effectiveness.}
    \label{fig:imperfect_att}
    \vspace{-1.5em}
\end{figure}

During benign operation, the defense need not execute the ensemble continuously.  AdROD-II triggers the ensemble defense only when an attack becomes effective. The system detects such cases via the inherent brittleness of adversarial patterns. In dynamic environments, changes in distance and perspective can hinder sustained patch effectiveness across the field of view~\cite{han2024visionguard}. As demonstrated in Fig.~\ref{fig:imperfect_att}, when subjected to physical attacks (SysAdv~\cite{wang2023does}, FTE~\cite{FTE}, and RP2~\cite{RP_2_on_detector}), an undefended YOLO detector consistently tracks the adversarially patched stop sign at a distance. Adversarial suppression takes effect after the vehicle is within a certain distance from the patch, and even then, the suppression is rarely absolute, still permitting intermittent successful detections. In practice, uncontrolled environmental variables, such as illumination changes and camera distortions, further degrade patch color fidelity and contrast, rendering continuous adversarial suppression practically unattainable.


AdROD-II exploits the kinematic discontinuity of attacked objects to detect adversarial interference. Specifically, once a vehicle enters the effective range of the patch, the detection trace of the attacked object exhibits a sustained and sudden loss of detection. In contrast, we observe that benign object disappearance typically occurs gradually: the bounding box shrinks as the object recedes, while occlusion or exiting the field of view progressively reduces its visible extent before disappearance. While benign FNs can also induce abrupt disappearances, such fluctuations are generally transient.


\textbf{Kinematic tracking and defense activation.} AdROD-II employs a Kalman Filter (KF) to track each object~\cite{luo2021multiple}. The KF takes detection results from the base detector to maintain a kinematic state $ [x, y, w, h, \dot{x}, \dot{y}, \dot{w}, \dot{h}]$ for each candidate object, where \((x, y)\) and \((w, h)\) denote the center coordinates and bounding box dimensions, respectively. A disappearance event is classified as \textit{abrupt} if the object vanishes without exhibiting the expected gradual scale reduction according to $\dot{w}$ and $\dot{h}$ typically observed during receding occlusion. Such objects are added to a {\em watch set} $\mathbb{S}$, where their states continue to be extrapolated by the KF for an extrapolation window of $\ell$ frames despite the lack of new observations. If the base detector re-detects and associates a detection with an object in $\mathbb{S}$ within this window, the disappearance is dismissed as a benign fluctuation, and the object is restored to standard tracking. However, if the disappearance persists beyond $\ell$ frames, AdROD-II identifies a sustained adversarial suppression and activates the HyperNetwork-based defense. We empirically set the extrapolation window $\ell=3$ based on a sensitivity analysis in \textsection\ref{sec:sensitive}.

\textbf{Sequential execution and early termination.} Once the ensemble defense is activated, AdROD-II sequentially executes the generated models and applies early termination to minimize computational load. For each model, the system retains only detections that can be associated with the KF-extrapolated states in the watch set $\mathbb{S}$. These validated recoveries are then merged into the final output according to the resolution logic defined in \textsection\ref{sec:rod}. AdROD-II halts further model execution once all watched objects in $\mathbb{S}$ are recovered for the current frame. Consequently, the number of executed models, denoted by $k$, adapts to the scene, reducing redundant computation. If no watched object can be recovered for a sustained duration, the system issues a defense-failure warning and deactivates the ensemble to conserve resources. Such a warning may indicate either successful adversarial suppression or a false activation. 

\textbf{Scope of activation.} AdROD-II activates only after an object has been detected and tracked at least once. This track may be established when the object first enters the field of view or later during approach, since physical attacks are often imperfect and may permit intermittent detections before sustained suppression. Once a tracked object subsequently disappears abruptly, AdROD-II activates the ensemble. False activations may arise from benign detector fluctuations, including transient FNs of genuine objects and the disappearance of short-lived FP tracks. For a transient FN, subsequent re-detection restores the genuine track and terminates ensemble execution. For a short-lived FP track, the ensemble typically fails to recover a corresponding object; persistent recovery failure eventually triggers the defense-failure warning and deactivation described above. 

\section{Implementation and Evaluation}
\label{evaluation}

\subsection{Implementation and Evaluation Methodology}
\label{implementation}

\textbf{Base detectors.}
We adopt a lightweight YOLO-small~\cite{jocher2020yolov5} as the primary base detector, and a two-stage Faster R-CNN~\cite{faster_rcnn} with a MobileNetV3 backbone to demonstrate architectural generalizability. We initialize both models with weights pretrained on the Microsoft COCO dataset~\cite{lin2014microsoft}.

\textbf{Implementation details.}   (1) \textit{Training:} We configure the low-rank HyperNetworks to update the base detector's backbone and neck modules, with rank $r\!=\!4$, block size $P\!=\!4$, and coefficient $\beta\!=\!1$. We use the COCO dataset ($640 \times 640$) for training, with a training period of 9 epochs, a batch size of 32, a training ensemble size of 3, and a learning rate controlled by cosine annealing ($10^{-4}$ to $10^{-5}$). The training takes about 10 hours on a Quadro 8000 GPU.
(2) \emph{Serving:} Since each detector is independently generated from sampled random noise, the ensemble size used to compute the diversity loss during training does not need to match the ensemble size used for serving. For AdROD-I, we set a serving ensemble size to $K\!=\!10$, $\tau_c=0.5$, and $Q=10$; for AdROD-II, we set maximum ensemble size to $K\!=\!10$ and extrapolation window $\ell=3$. \textsection\ref{sec:sensitive} provides a sensitivity analysis on these settings.

\textbf{Baselines.} 
We benchmark AdROD against: an undefended Vanilla model, adversarial training (AdvTrain), and five representative defenses designed without considering the tested attacks, including JPEG~\cite{guo2017countering}, LGS~\cite{LGS}, SAC~\cite{liu2022SAC}, Jedi~\cite{tarchoun2023jedi}, and ObjectSeeker~\cite{Objseeker}, using their original configurations. For AdvTrain, we adopt a naming convention of \textit{AdvTrain-Class} to denote a model that is adversarially trained against a specific class of objects instrumented with regional patch attacks.

\textbf{Metrics.} We evaluate AdROD using the following three metrics: (1) {\em ASR} as defined in \textsection\ref{approach}; (2) \emph{benign detection accuracy} measured by standard object detection metrics including precision, recall, and mAP. We report mAP at IoU 0.5 (mAP50) and the average mAP across thresholds from 0.5 to 0.95 (mAP50:95) on the COCO validation dataset; and (3) \emph{runtime overhead} measured in the unit of $\lambda$, where $\lambda$ denotes the per-image inference latency of the vanilla YOLO detector (including NMS). Expressing overhead in $\lambda$ ensures a device-agnostic benchmark for computational cost.


\subsection{Benign Detection Accuracy and Overhead}
\label{sec:acc and overhead}

\textbf{Benign detection accuracy.} Table~\ref{tab:clean_acc} summarizes the results on the COCO validation dataset. AdROD-II is excluded because COCO lacks the continuous video frames required for its tracking mechanism. Jedi is omitted because only a MATLAB implementation is available, and a direct overhead comparison would be unfair. From Table~\ref{tab:clean_acc}, AdROD-I's benign detection accuracy is comparable to the vanilla model and superior to adversarial training baselines. 

\begin{figure}[!t]
\centering
\includegraphics[width=1\linewidth]{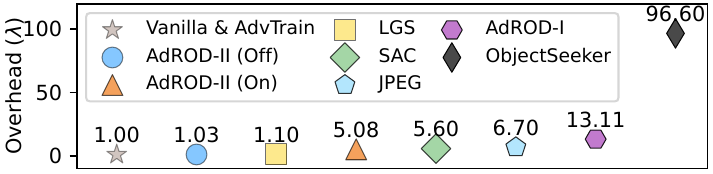}
\caption{Computational overhead of defenses. The execution time ratio is normalized relative to the vanilla model, with lower values indicating lower overhead.}
\label{fig:overhead}
\end{figure}

\begin{table}[!t]
\centering
\caption{Benign detection accuracy.}
\label{tab:clean_acc}
\begin{threeparttable}
\setlength{\tabcolsep}{3pt}
\resizebox{0.99\linewidth}{!}{
\renewcommand{\arraystretch}{1}
\begin{tabular}{lccccc}
\toprule
\textbf{Method} &\textbf{Dep.} & \textbf{Precision} & \textbf{Recall} & \textbf{mAP50} & \textbf{mAP50:95} \\ 
\midrule
Vanilla                       & --        & $0.666$ & $0.515$ & $0.562$ & $0.371$ \\
LGS \cite{LGS}                & --        & $0.613$ & $0.436$ & $0.472$ & $0.301$ \\
JPEG \cite{guo2017countering} & --        & $0.656$ & $0.492$ & $0.537$ & $0.348$ \\

ObjectSeeker \cite{Objseeker} & --        & $0.662$ & $0.520$ & $0.557$ & $0.369$ \\ 
SAC \cite{liu2022SAC}         & $\circ$   & $0.667$ & $0.513$ & $0.558$ & $0.369$ \\ 
AdvTrain-Person               & $\bullet$ & $0.642$ & $0.499$ & $0.539$ & $0.344$ \\
AdvTrain-Stopsign             & $\bullet$ & $0.634$ & $0.503$ & $0.539$ & $0.346$ \\
\midrule
\textbf{AdROD-I}              & --        & $\mathbf{0.646}$ & $\mathbf{0.508}$ & $\mathbf{0.541}$ & $\mathbf{0.358}$ \\
\bottomrule
\end{tabular}
}
\begin{tablenotes}
\scriptsize
\item[] \parbox[t]{1.00\linewidth}{Dep. indicates training-time attack dependency: $\circ$ denotes attack-pattern assumptions or patch-specific training; $\bullet$ denotes attack-specific adversarial training; -- denotes no attack-related training.}
\end{tablenotes}
\end{threeparttable}
\end{table}

\begin{table}[!t]
\centering
\caption{Overhead breakdown of AdROD-I and AdROD-II.}
\label{tab:rod_overhead_breakdown}
\setlength{\tabcolsep}{3pt} 
\resizebox{0.99\linewidth}{!}{
\renewcommand{\arraystretch}{1}
\begin{tabular}{lccc}
\toprule
 & \multicolumn{3}{c}{\textbf{Overhead ($\times \lambda$)}} \\
\cmidrule(lr){2-4}
\textbf{Component} & \textbf{AdROD-I} & \textbf{AdROD-II (On)} & \textbf{AdROD-II (Off)} \\ 
\midrule
Weight generation & $0.169K$ & $0.169k$ & -- \\ 
Input transform & $0.041K$ & $0.041k$ & -- \\
Ensemble inference & $0.938K$ & $1.000k$ & -- \\
Anomaly detection & -- & $0.03$ & $0.03$ \\
Result merge & $1.63$ & $1.00 $ & $1.00$ \\ 
\midrule 
\textbf{Total} & $\mathbf{1.148K \!+\! 1.63}$ & $\mathbf{1.21k \!+\! 1.03}$ & $\mathbf{1.03}$ \\ 
\bottomrule
\end{tabular}
}
\begin{tablenotes}
\scriptsize
\item[] \parbox[t]{1.00\linewidth}{Note: Ensemble member inference in AdROD-I excludes NMS, whereas AdROD-II uses the full detection pipeline; $k \le K$; ``On''/``Off'' indicates ensemble activation status.}
\end{tablenotes}
\vspace{-1em}
\end{table}

\textbf{Baseline overhead.} Fig.~\ref{fig:overhead} presents computational overhead in $\lambda$. AdvTrain introduces no additional inference latency, but fails to generalize to unseen threats as shown in \textsection\ref{sec:syn} shortly. While LGS is fast ($1.1\lambda$), it sacrifices accuracy due to global image distortion. Other baselines introduce significant overhead: SAC relies on a U-Net backbone ($5.6\lambda$), JPEG incurs substantial CPU encoding costs ($6.7\lambda$), and ObjectSeeker delivers the highest latency ($96.6\lambda$) due to exhaustive sliding-mask passes with numerous detector inferences.

\textbf{AdROD overhead breakdown.} Table~\ref{tab:rod_overhead_breakdown} breaks down the overhead of the two AdROD variants. Although AdROD-I ($K\!=\!10$) achieves strong robustness (see \textsection\ref{sec:syn}), its continuous ensemble execution incurs a $13.11\lambda$ static overhead, limiting throughput to 5.4 FPS on an NVIDIA Jetson AGX Xavier. AdROD-II bypasses this bottleneck via an on-demand policy. In attack-free scenarios, it disables the ensemble and runs only the anomaly detector ($1.03\lambda$), achieving 68.7 FPS. When attacked, the activated ensemble executes sequentially and terminates early, thereby dynamically adjusting the subset size $k$ ($k \le K$). Consequently, even though the per-model overhead coefficient in AdROD-II is marginally higher, AdROD-II's aggregate overhead reduces to an average of  $5.08\lambda$ (13.9 FPS), as detailed in \textsection\ref{sec:outdor_eval} shortly. Furthermore, AdROD-II's average throughput of 13.9 FPS satisfies the representative 100-ms perception deadline reported in~\cite{lin2018architectural}.


\subsection{Sensitivity Analysis and Ablation Study}
\label{sec:sensitive}

\begin{figure}[!t]
    \centering
    \setlength{\tabcolsep}{1pt} 
    \renewcommand{\arraystretch}{0.95} 
    \begin{tabular}{ccc}
        \includegraphics[height=2.5cm]{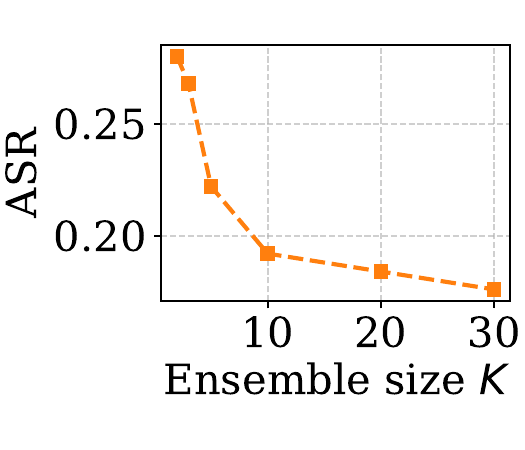} &
        \includegraphics[height=2.5cm]{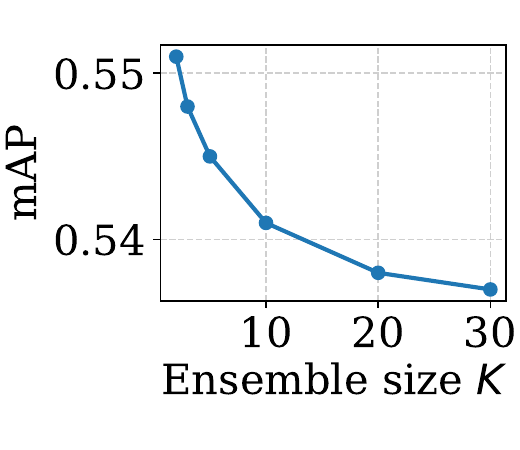} &
        \includegraphics[height=2.5cm]{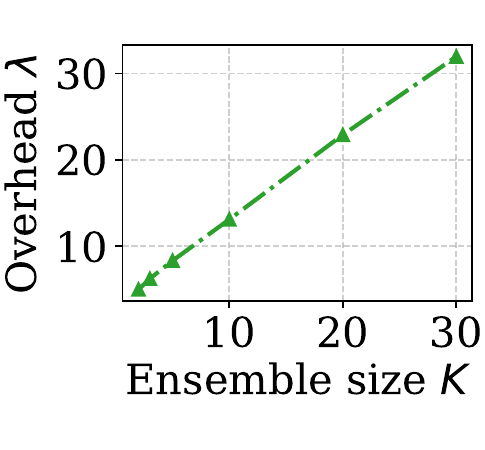} \\[-.5em]
        {\small \textbf{(a) Robustness vs. $K$}} & 
        {\small \textbf{(b) Accuracy vs. $K$}} & 
        {\small \textbf{(c) Overhead vs. $K$}} \\ [5pt]
        \multicolumn{3}{c}{
            \begin{tabular}{cc}
                \includegraphics[width=0.45\linewidth]{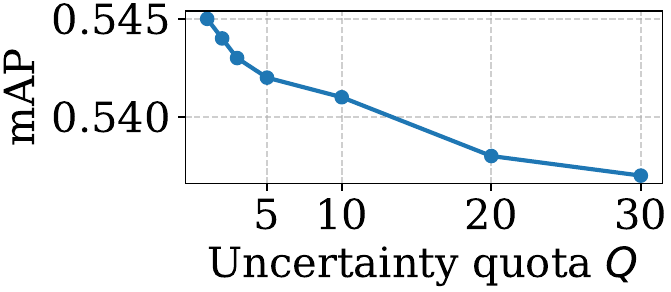} &
                \includegraphics[width=0.45\linewidth]{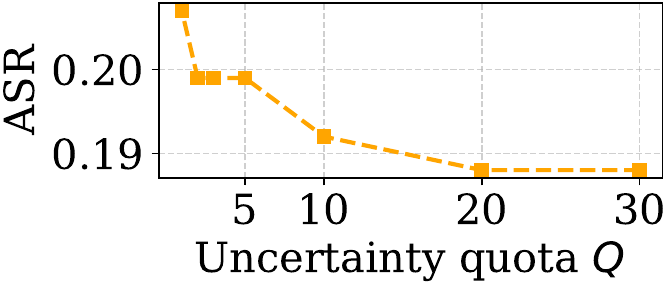} \\ 
                {\small \textbf{(d) Accuracy vs. $Q$}} & 
                {\small \textbf{(e) Robustness vs. $Q$}} 
            \end{tabular}
        }
    \end{tabular}
    \caption{(a–c) illustrate how increasing $K$ affects accuracy, robustness, and runtime overhead, while (d–e) show how varying $Q$ influences accuracy and robustness in AdROD-I.}
    \label{fig:k_and_kvar}
\end{figure}

\textbf{Impact of ensemble size $K$ on AdROD-I.} 
Fig.~\ref{fig:k_and_kvar}(a–c) shows the effect of varying $K \in \{2,3,4,10,20,30\}$. ASR decreases with increasing ensemble size, but further gains diminish beyond $K\!=\!10$. Conversely, benign detection accuracy (mAP) shows a slight decline as the larger ensemble increases the statistical probability of introducing FPs. Meanwhile, computational overhead grows nearly linearly with $K$. Thus, selecting an appropriate $K$ is essential. 


\textbf{Impact of uncertainty quota $Q$ on AdROD-I.} 
The parameter $Q$ controls the number of highly uncertain bounding boxes retained during the ensemble merging process. As shown in Fig.~\ref{fig:k_and_kvar}(d–e), increasing $Q$ effectively lowers the ASR but slightly degrades mAP. This suggests a trade-off between precision and defense strength. Because overall ASR and mAP are less sensitive to $Q$ than to $K$, $Q$ can be employed as the secondary tuning parameter.

\textbf{Impact of extrapolation window $\ell$ on AdROD-II.} 
The parameter $\ell$ controls the sensitivity of the defense trigger. Because evaluating $\ell$ requires continuous video frames, we conduct this analysis using a physical outdoor testbed, which will be detailed in \textsection\ref{sec:outdor_eval} shortly.
An aggressive setting of $\ell\!=\!1$ leads to a marginal robustness gain (ASR $0.033$ vs. $0.058$ at $\ell\!=\!3$). However, it leads to high overhead, because the occasional benign FNs produced by the base detector trigger the ensemble defense at almost every momentary loss of an object. Conversely, a highly tolerant setting of $\ell\!=\!5$, which leads to an ASR of $0.079$, introduces extra delays before classifying sustained disappearances as attacks, thereby increasing ASR undesirably. Consequently, $\ell\!=\!3$ strikes a satisfactory balance, leveraging kinematic memory to efficiently bridge benign FNs while ensuring prompt defense activation.

\begin{table}[t]
\centering
\caption{Ablation study on weight and functional diversity against stop-sign-targeted SysAdv attack. Affirmative fusion, which retains any detection, is applied for the best robustness.}
\label{tab:diversity_contribution}
\setlength{\tabcolsep}{3pt}
\renewcommand{\arraystretch}{0.99}
\begin{tabular}{lcc}
\toprule
\textbf{Defense Variant} & \textbf{ASR}  & \textbf{mAP50}  \\
\midrule
\multicolumn{3}{l}{\textit{Training-free Baselines}} \\
\quad Vanilla & $0.728$ & $0.562$  \\
\quad Vanilla + Image transform  & $0.834$ & $0.132$ \\
\midrule
\multicolumn{3}{l}{\textit{Weight and Functional Diversity}} \\
\quad None                       & $0.705$ & $0.562$ \\
\quad Functional diversity only ($P{=}4$)  & $0.498$ & $0.476$ \\
\quad Weight diversity only ($\beta{=}1$) & $0.613$ & $0.544$ \\
\quad Weight + Functional diversity ($\beta{=}1$, $P{=}4$) & $0.195$ & $0.424$ \\
\bottomrule
\end{tabular}
\vspace{-1em}
\end{table}

\textbf{Ablation.} In Table~\ref{tab:diversity_contribution}, we isolate the contributions of each diversity component. First, we examine whether block-wise pixel shuffling, applied without retraining, provides robustness gains. As shown under \textit{Training-free Baselines}, its direct application degrades detection accuracy while increasing ASR, indicating that the transform alone is ineffective as a defense. Then, we evaluate each diversity component by training the low-rank HyperNetworks under various configurations. Functional diversity and weight diversity each improve robustness when applied individually, but neither is sufficient to provide strong protection. When combined, the two components achieve substantially stronger robustness, demonstrating their complementary effects. Although this joint configuration reduces the mAP, AdROD-I recovers most of the performance loss, as seen in Table~\ref{tab:clean_acc}. Furthermore, AdROD-II avoids this degradation by selectively fusing only validated and recovered objects with the base detector outputs, thereby preserving accuracy on benign samples. Overall, these results show that both weight diversity and functional diversity are essential for effective defense in object detection.

\subsection{Robustness Analysis at the Perception Level}
\label{section: eval_model_robustness}

\subsubsection{Physical Outdoor Evaluation}
\label{sec:outdor_eval}

\begin{figure}[t]
  \centering
\includegraphics[width=\linewidth]{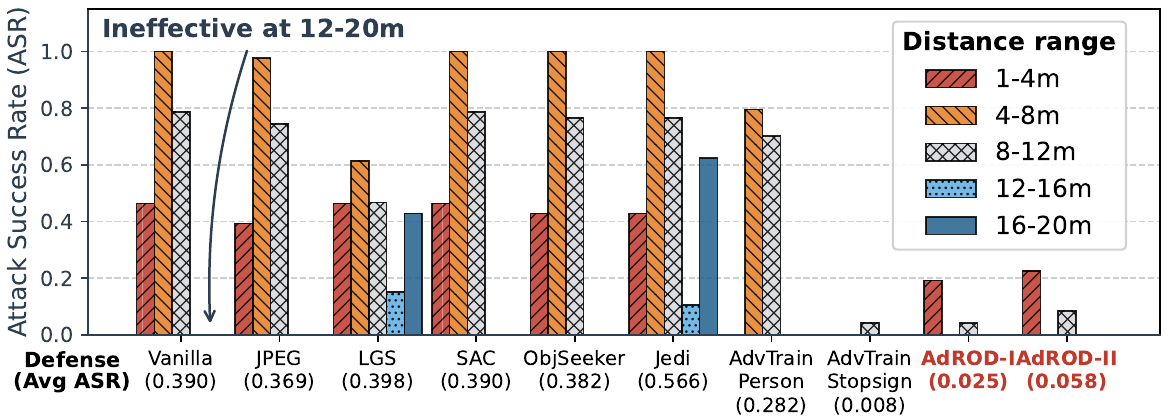}
 \vspace{-1.5em}
  \caption{ASR of defense methods against SysAdv attack.}
  \label{fig:defense_effectiveness}
  \vspace{-1em}
\end{figure}

We evaluate AdROD-I and AdROD-II on an outdoor testbed using a physically deployed SysAdv~\cite{wang2023does} adversarial patch, which is selected due to its strong digital-to-physical transferability and system-level effectiveness. The patch is printed and attached to a stop sign placed on a restricted private road. A victim camera mounted on a vehicle approaches the stop sign from a distance of 20~m. Fig.~\ref{fig:defense_effectiveness} shows that on the vanilla model, the patch is ineffective at 20-12 m because its adversarial patterns cannot be sufficiently captured by the camera, reaches ASR 0.78 at 12-8 m and 1.00 at 8-4 m, then falls to 0.46 at 4-1 m because such close-ups are rare in patch optimization.


\textbf{Effectiveness.}
Fig.~\ref{fig:defense_effectiveness} also compares the robustness of different defense methods. Both AdROD variants significantly reduce the ASR across all distance ranges. AdROD-I achieves the lowest average ASR of 0.025, while AdROD-II achieves an average ASR of 0.058. The slightly higher ASR of AdROD-II is expected because its defense is triggered only after the disappearance of the victim object is detected. Despite this delayed activation, AdROD-II remains the second-best attack-agnostic defense in this experiment. 
Moreover, AdROD achieves performance comparable to \emph{AdvTrain-StopSign}, which is trained with attack-specific knowledge, and substantially outperforms \emph{AdvTrain-Person}, which is trained on a different object class. These results further demonstrate the superior generalization ability of AdROD.

\begin{figure}[t]
    \centering
    \begin{subfigure}[b]{1\linewidth}
        \includegraphics[width=1\linewidth]{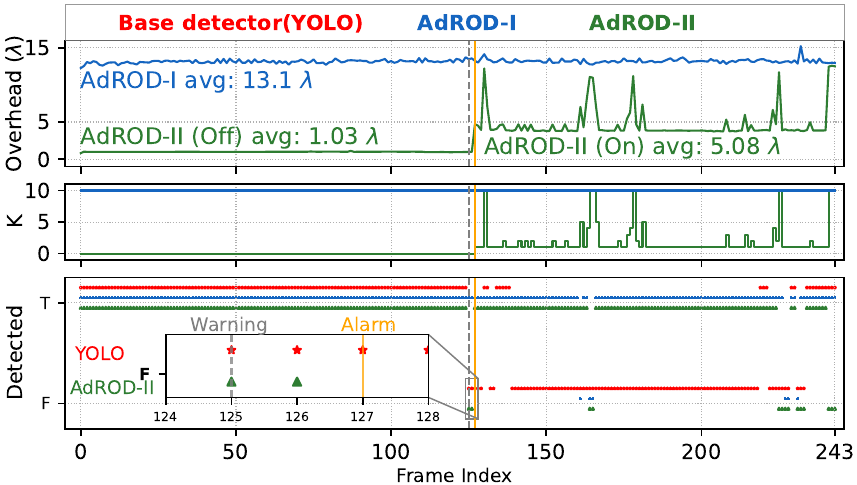}
        \caption{Runtime behaviors of vanilla YOLO, AdROD-I, and AdROD-II. `T' indicates recovered stop sign detection; `F' indicates miss.}
        \label{runtime_a}
    \end{subfigure}%
    \\ 
    \vspace{0em}
    \begin{subfigure}[b]{1\linewidth}
    \centering
        \includegraphics[width=1\linewidth]{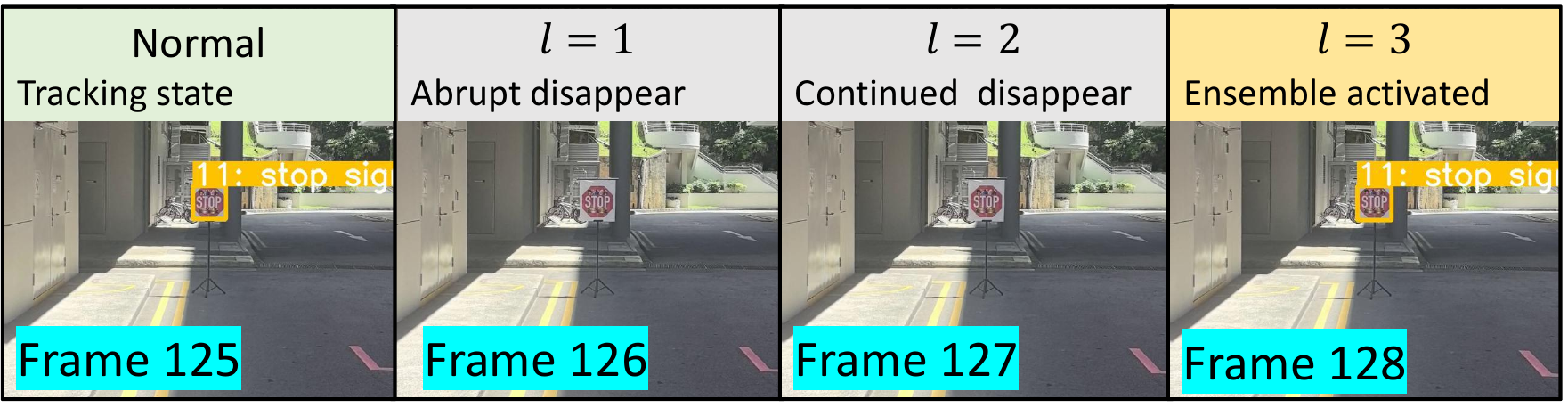}
        \caption{Operational states of AdROD-II (Frames 125–128): transition from tracking to ensemble activation.}
        \label{runtime_b}
    \end{subfigure}%
    \caption{Physical experiment against SysAdv. The attack becomes effective from Frame 126 to the vanilla YOLO.}
    \vspace{-1em}
    \label{fig:realtime}
\end{figure}

\textbf{Runtime dynamics.} Fig.~\ref{runtime_a} illustrates the execution trace of AdROD compared with the vanilla model, covering runtime overhead, ensemble size, and object recovery status. Before Frame 126, when the camera remains distant and the adversarial patch has no effect, AdROD-II maintains a near-baseline overhead of $1.03\lambda$. As the attack manifests at Frame 126, AdROD-II leverages KF-extrapolated states to flag the anomaly and transitions the lost object to its watch set $\mathbb{S}$. By Frame 128, after three consecutive misses, the ensemble defense is activated with sequential ensemble execution and early termination to bound overhead. Fig.~\ref{runtime_b} visualizes this state transition: from normal tracking (Frame 125), to watching (Frames 126–127), and finally to defense activation (Frame 128). Together, these results show how AdROD-II enables robust recovery with minimal ensemble execution at runtime.

\subsubsection{Real-world Third-party Vehicle Evaluation}
\label{subsec:competition}

\begin{figure}[t]
\centering
\includegraphics[width=1\linewidth]{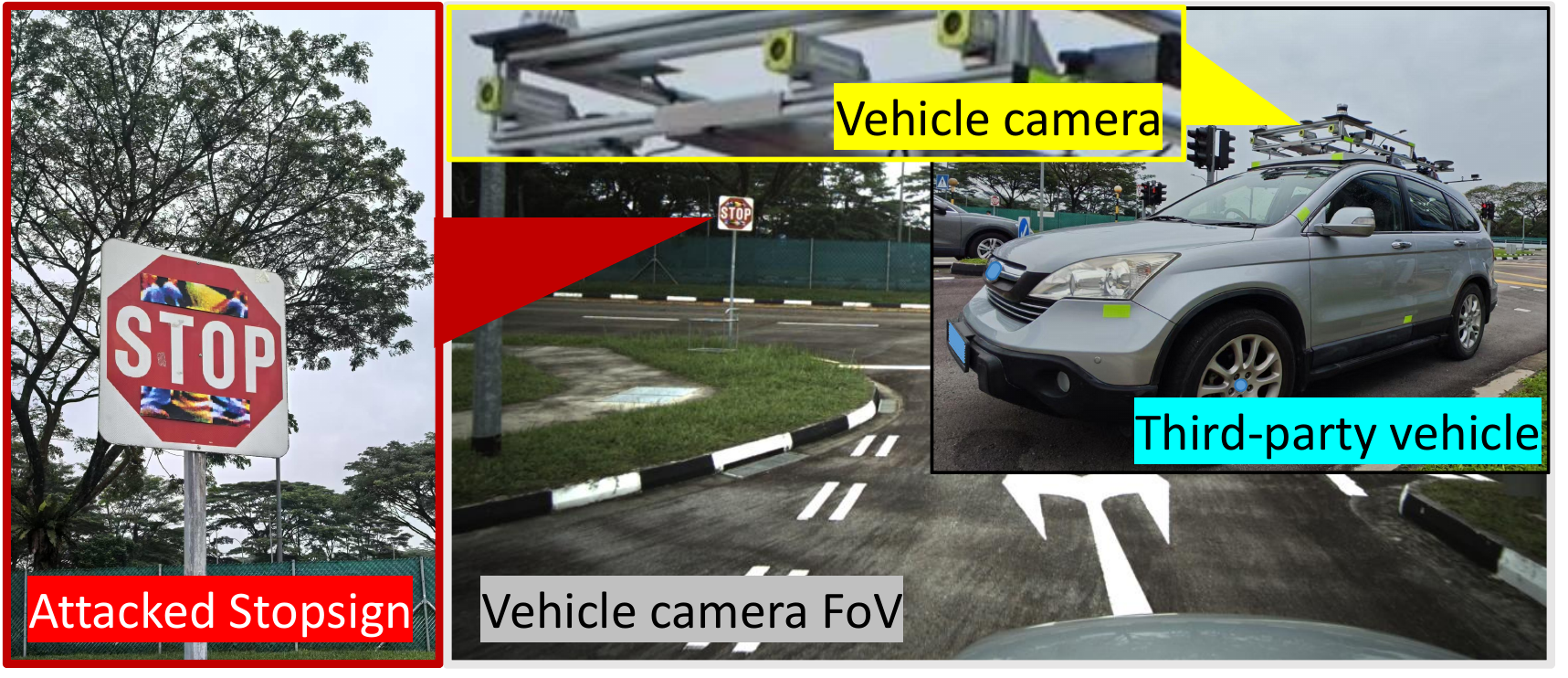}
\caption{Experiment in the vehicle security challenge.}
\label{fig:physical_trail}
\vspace{-1em}
\end{figure}

In a third-party vehicle security challenge, an Allied Vision Alvium G1-510C camera recorded a vehicle traveling at 20 km/h and 10 FPS. Its wide field of view (74.5° horizontal and 54° vertical) made the sign smaller than in \textsection\ref{sec:outdor_eval}, so we optimized SysAdv at 640×640 and 1280×1280 (LowRes and HighRes). Recorded data were released to us after collection. Fig.~\ref{fig:physical_trail} illustrates the adversarial setup used for evaluation.


\emph{SysAdv-HighRes} achieves an ASR of 0.759 on the vanilla model, which decreases to 0.310 and 0.345 with AdROD-I and AdROD-II, respectively. \emph{SysAdv-LowRes} achieves an ASR of 0.719 on the vanilla model, which decreases to 0.219 and 0.344 with AdROD-I and AdROD-II, respectively. These results demonstrate AdROD's effectiveness and provide evidence of transfer across the evaluated camera setups and real-world conditions. Prior system-level evaluation of stop-sign attacks~\cite{wang2023does} suggests that traffic-rule violations require sustained detection failures over multiple frames. Therefore, \textsection\ref{sec:e2e} further evaluates whether AdROD preserves system-level safety in the full perception--control loop.

\begin{figure*}[!t]
    \centering
    \setlength{\tabcolsep}{1pt} 
    \footnotesize 
    \resizebox{1.0\linewidth}{!}{%
    \begin{tabular}{cccc} 
        {\includegraphics[width=0.12\linewidth]{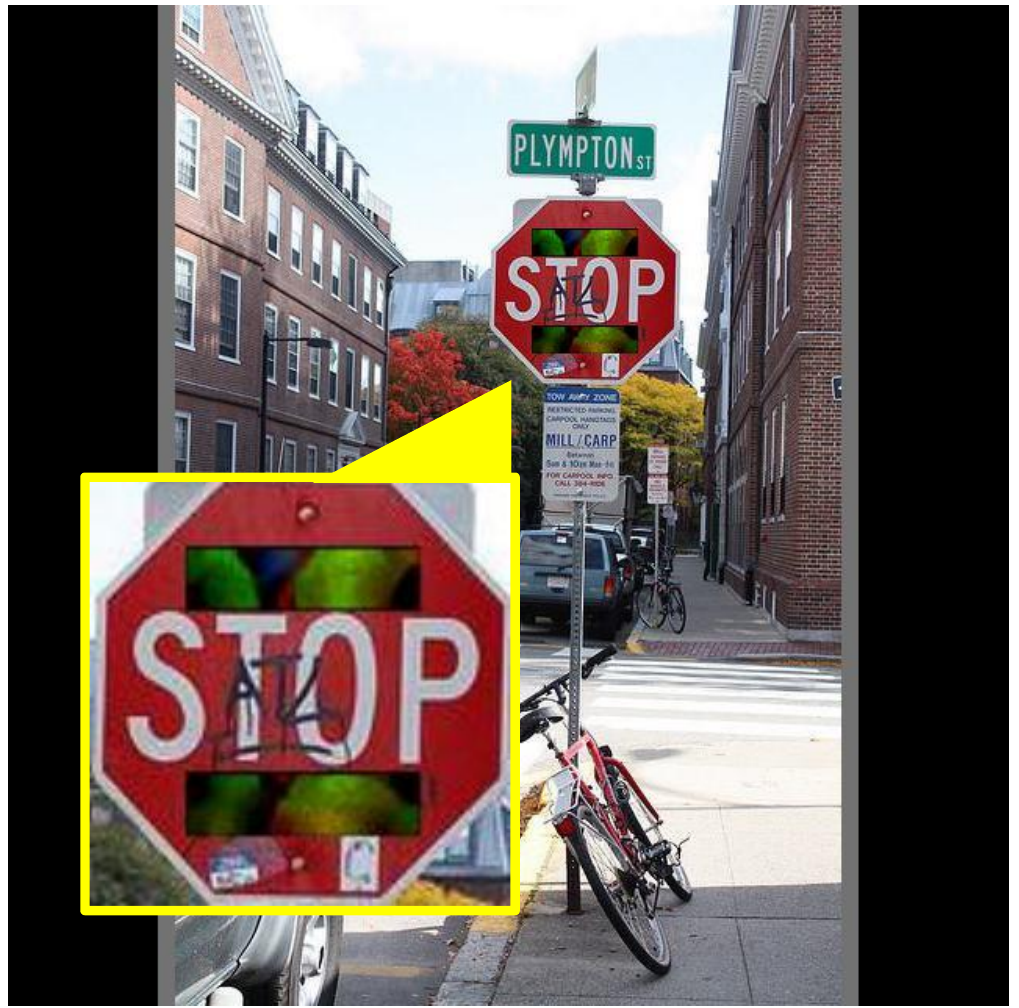}}  
        {\includegraphics[width=0.12\linewidth]{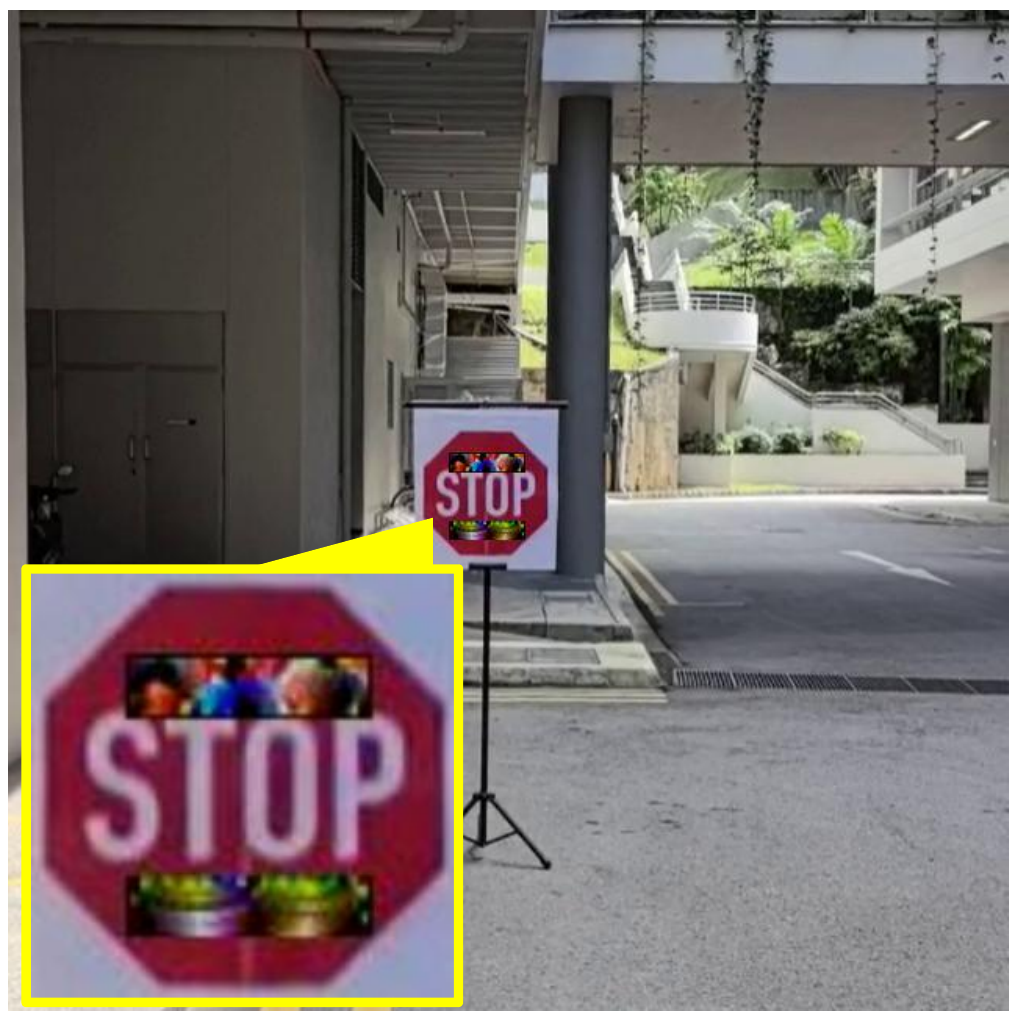}} &
        {\includegraphics[width=0.12\linewidth]{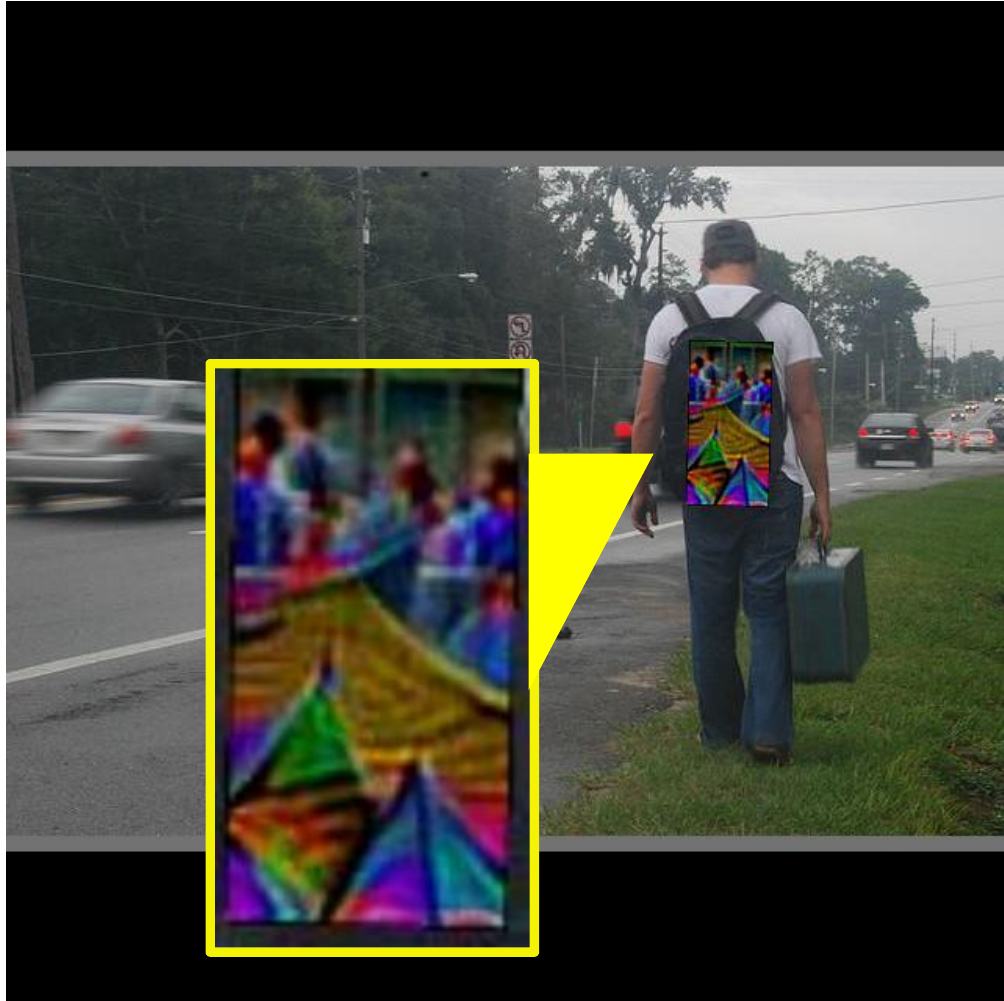}}   
        {\includegraphics[width=0.12\linewidth]{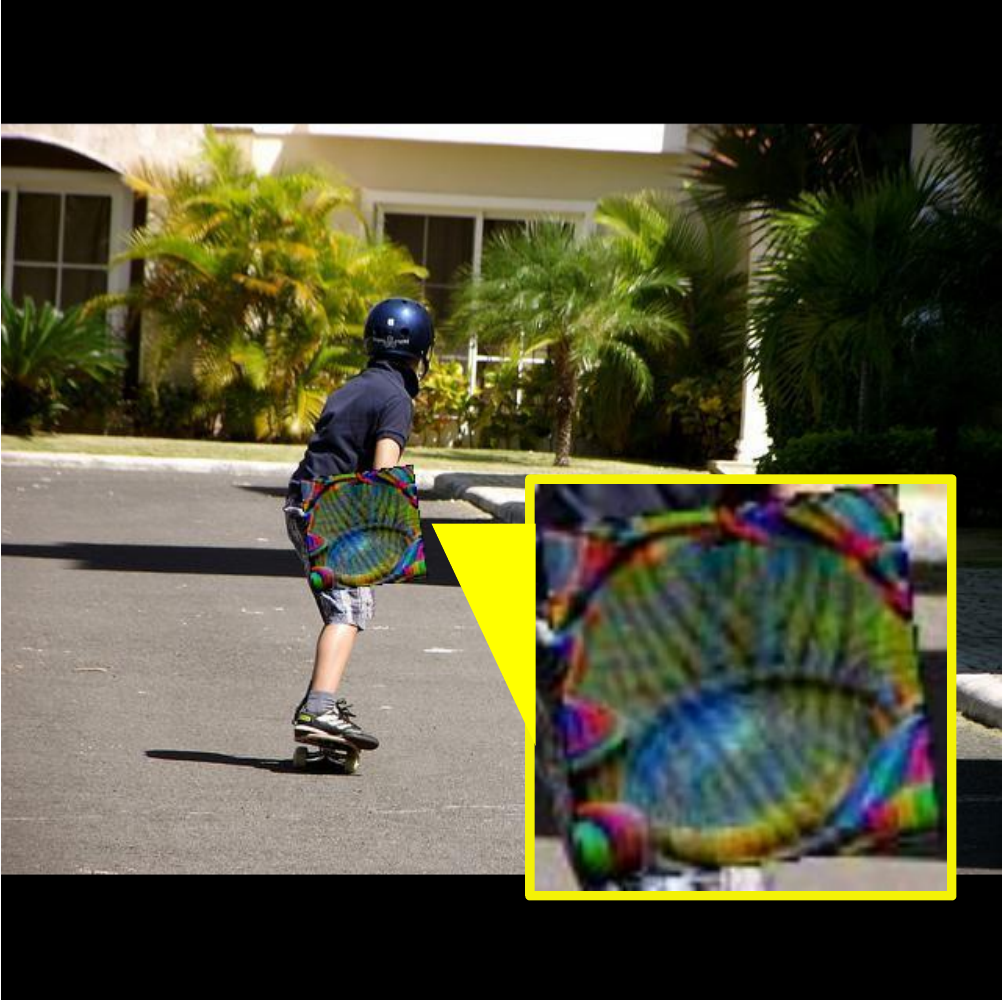}}  & 
        {\includegraphics[width=0.12\linewidth]{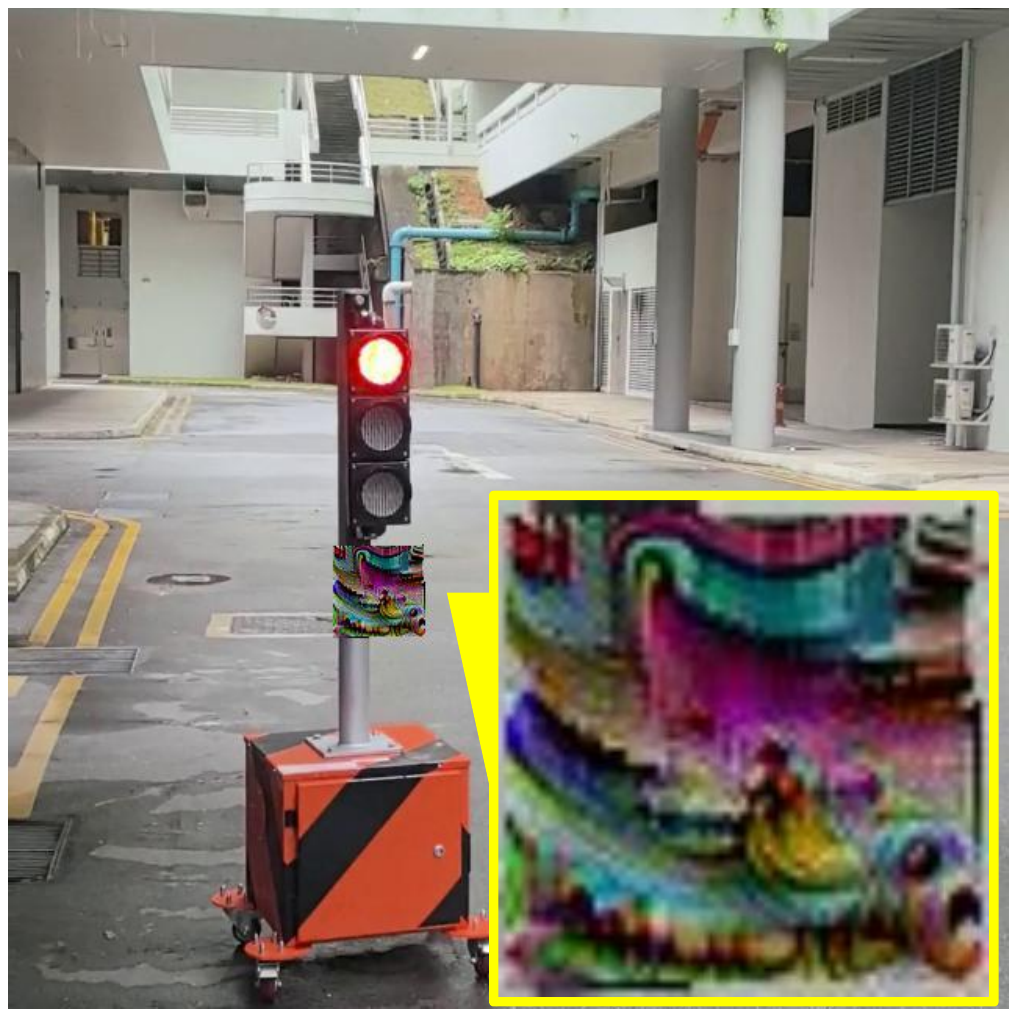}} 
        {\includegraphics[width=0.12\linewidth]{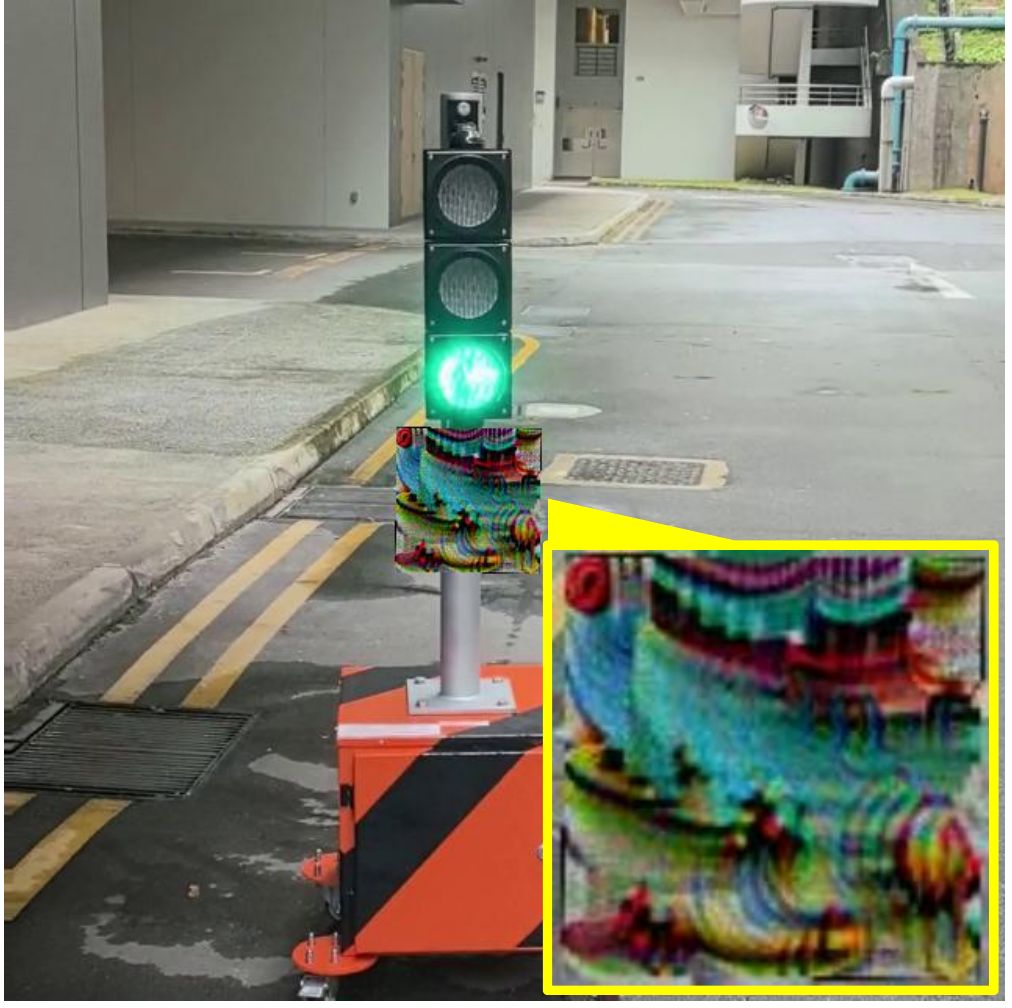}} &
        {\includegraphics[width=0.12\linewidth]{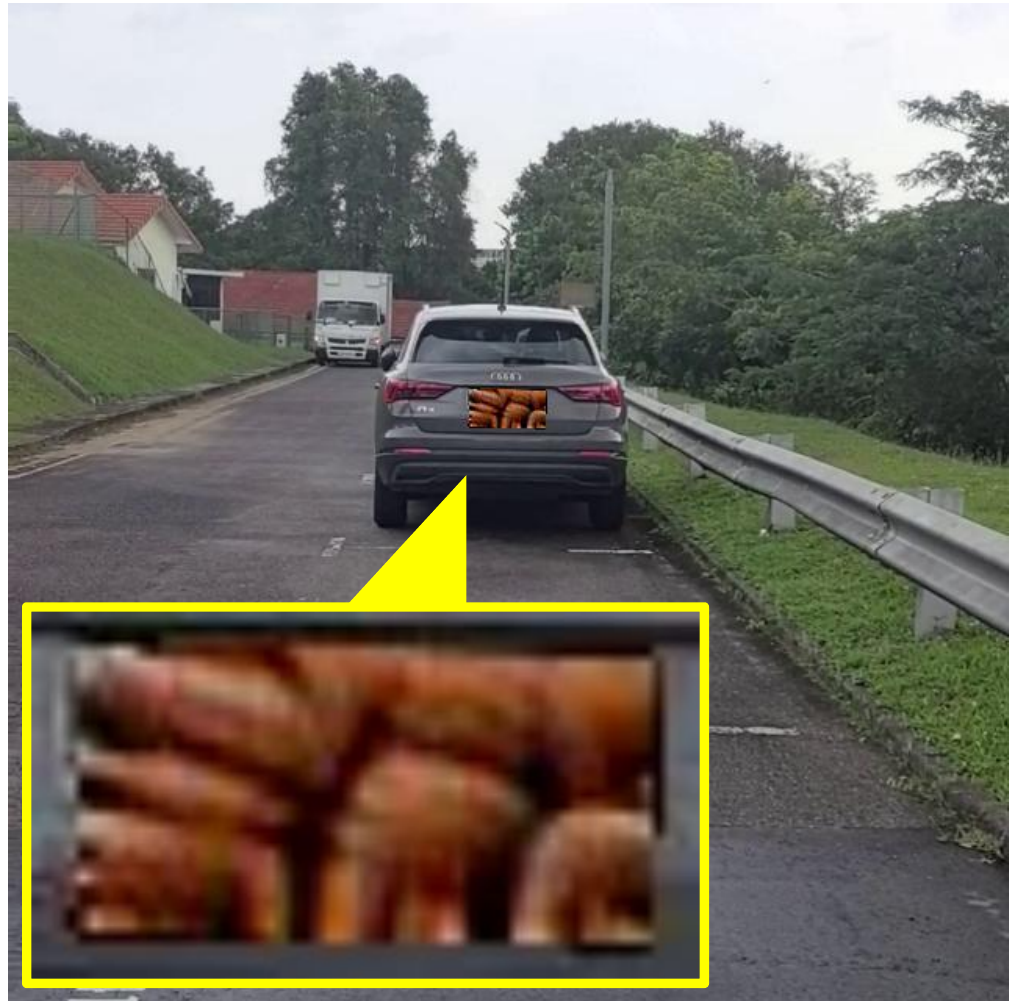}}   
        {\includegraphics[width=0.12\linewidth]{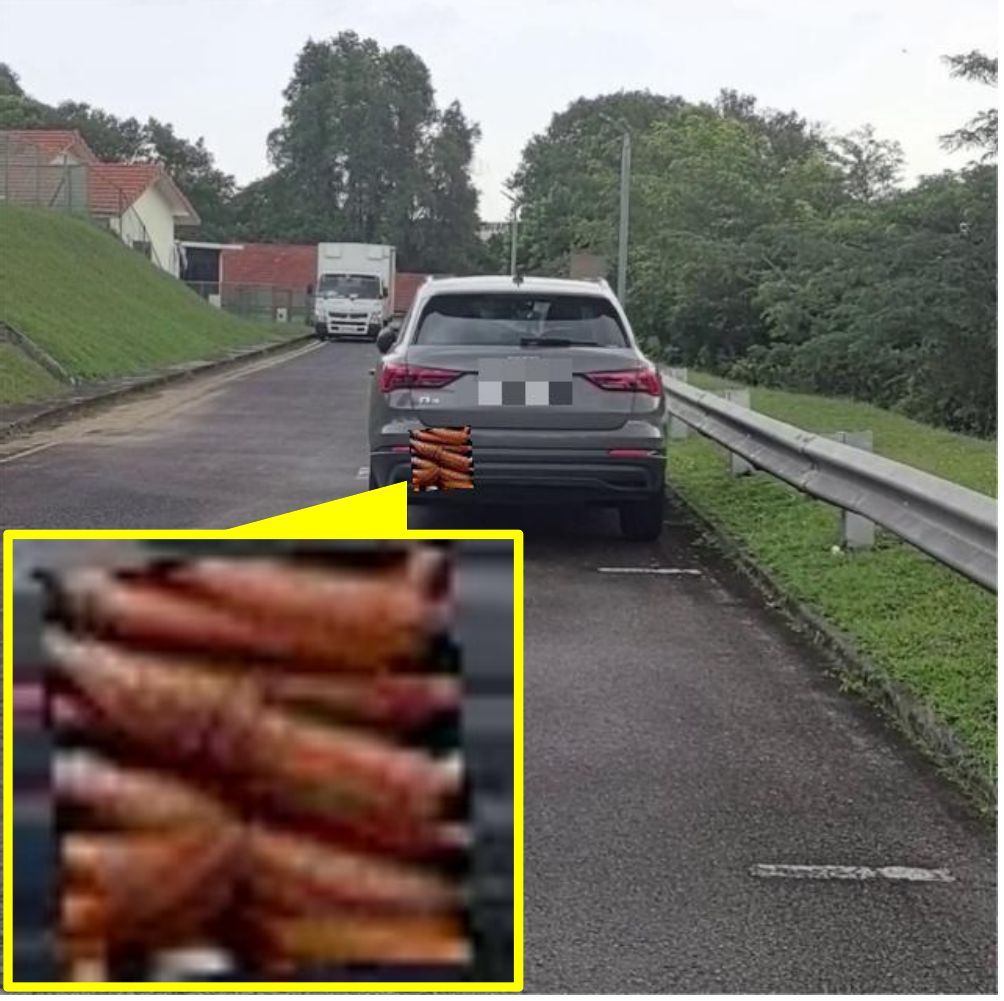}}   
        \\
         \multicolumn{1}{l}{\hspace{0.4em} SysAdv-smooth \hspace{2.5em}  SysAdv} &  \multicolumn{1}{l}{\hspace{2em} Rectangle  \hspace{3.5em}  Square} & 
         \multicolumn{1}{l}{\hspace{2.5em} Flush  \hspace{5.5em}  Wide} & \multicolumn{1}{l}{\hspace{2.5em} Fixed   \hspace{5.5em} 
 Free} \\
        \textbf{(a) Stop sign} (COCO \& self-collected) &  \textbf{(b) Person} (COCO) & \textbf{(c) Traffic light} (self-collected) & \textbf{(d) Car} (self-collected)
        
    \end{tabular}
    }
    \caption{Adversarial patch attacks with different attack targets and characteristics. The callouts show zoomed-in patches.}
    \label{fig:attack_examples}
    \vspace{-1em}
\end{figure*}

\begin{table*}
\centering
\caption{ASR of attacks against baseline defense methods and AdROD-I.}
\vspace{-.5em}
\resizebox{1.0\linewidth}{!}{%
\begin{tabular}{ccccccccccc}
\toprule

 & & & \textbf{No Defense} & \multicolumn{6}{c}{\textbf{Baseline Defenses}} & \textbf{Ours} \\
\cmidrule(lr){4-4} \cmidrule(lr){5-10} \cmidrule(lr){11-11}
\textbf{Object} & \textbf{Data} & \textbf{Attack Type} & Vanilla & JPEG~\cite{guo2017countering} & LGS~\cite{LGS} & SAC~\cite{liu2022SAC} & Jedi~\cite{tarchoun2023jedi} & ObjSeeker~\cite{Objseeker} & AdvTrain-Stopsign & AdROD-I \\
\toprule

\multirow{4}{*}{Stop sign}  & \multirow{4}{*}{COCO}
& $\text{RP}_2$  & $0.701$ & $0.682$ & $0.490$ & $0.701$ & $0.678$ & $0.602$ & $\underline{\mathbf{0.153}}$ & $\mathbf{0.180}$ \\
& & FTE & $0.851$ & $0.808$ & $0.644$ & $0.851$ & $0.789$ & $0.759$ & $\underline{\mathbf{0.187}}$ & $\mathbf{0.218}$ \\
& &SysAdv   & $0.728$ & $0.751$ & $0.521$ & $0.728$ & $0.655$ & $0.651$ & $\underline{\mathbf{0.190}}$ & $\mathbf{0.195}$ \\
& &SysAdv-smooth & $0.375$ & $0.360$ & $0.414$ & $0.375$ & $0.483$ & $0.287$ & $\underline{\mathbf{0.093}}$ & $\mathbf{0.153}$ \\
\midrule

\multirow{2}{*}{Person}  & \multirow{2}{*}{COCO}
& {Rectangle}&  $0.685$ & $0.674$ & $0.563$ & $0.635$ & $0.587$ & $\mathbf{0.524}$ & $0.676$ & $\underline{\mathbf{0.172}}$ \\
& &{Square}& $0.587$ & $0.567$ & $0.298$ & $\mathbf{0.124}$ & $0.332$ & $0.336$ & $0.646$ & $\underline{\mathbf{0.088}}$ \\
\midrule

\multirow{2}{*}{Traffic light} & \multirow{2}{*}{Self-collected}
& {Flush}& $0.443$ & $0.424$ & $0.167$ & $\underline{\mathbf{0.057}}$ & $0.261$ & $0.290$ & $0.234$ & $\mathbf{0.096}$ \\
& &{Wide}&  $0.729$ & $0.765$ & $0.351$ & $\underline{\mathbf{0.131}}$ & $0.547$ & $0.512$ & $0.483$ & $\mathbf{0.202}$\\
\midrule
\multirow{2}{*}{Car} & \multirow{2}{*}{Self-collected}
& {Fixed} & $0.913$ & $0.895$ & $0.648$ & $0.644$ & $0.882$ & $0.913$ & $\mathbf{0.614}$ & $\underline{\mathbf{0.247}}$ \\
& &{Free}& $0.868$ & $0.881$ & $0.667$ & $0.534$ & $0.873$ & $0.873$ & $\mathbf{0.504}$ & $\underline{\mathbf{0.288}}$ \\

\midrule
\multicolumn{3}{c}{\textbf{Benign mAP@50}}
& $0.562$
& $0.537$ 
& $0.472$
& $0.558$
& -- 
& $0.557$ 
& $0.539$ 
& $0.541$ \\
\bottomrule

\end{tabular}
}

\vspace{0.3em}
\footnotesize\textit{Note:} \underline{\textbf{Bold and underlined}} text indicates the lowest ASR (best), and \textbf{bold} text indicates the second lowest.  ``--'' indicates unavailable benign mAP.
\label{tab:compare_digital}
    \vspace{-1em}
\end{table*}

\subsubsection{Diverse Synthetic Evaluation}
\label{sec:syn}

We evaluate AdROD against a broad spectrum of physical-style adversarial attacks, varying in \textit{shape} (rectangular vs. square), \textit{size} (flush vs. wide), \textit{placement} (fixed vs. free), and \textit{target class}, as illustrated in Fig.~\ref{fig:attack_examples}. The evaluation is conducted on COCO validation subsets (256 stop signs, 426 persons) and a self-collected dataset comprising 244 frames of stop signs, 505 frames of traffic lights, and 218 frames of cars. Each frame contains a single attacked object with perturbations digitally applied. Because some datasets lack consecutive frames, this evaluation focuses on AdROD-I.

\textbf{Effectiveness.} Table~\ref{tab:compare_digital} shows that AdROD-I achieves the lowest or second-lowest ASR across all configurations. Conversely, baselines are configuration-sensitive. Specifically, \emph{AdvTrain-StopSign} degrades on unseen classes and yields high (>0.5) ASRs on cars. \emph{SAC}~\cite{liu2022SAC} performs well on the square-noise traffic-light attack but degrades on several other tested shapes/textures.
\emph{LGS}'s gradient-based filtering~\cite{LGS} and \emph{Jedi}'s entropy-based localization~\cite{tarchoun2023jedi} show degraded robustness against {\em SysAdv-smooth}'s smooth-textured patches. \emph{ObjectSeeker}~\cite{Objseeker} is less effective against adversarial patches that overlap the target object. Finally, \emph{JPEG} compression~\cite{guo2017countering}  is ineffective against visually prominent patches.



\begin{figure}[t]
    \centering
    \setlength{\tabcolsep}{0.5pt} 
    \begin{tabular}{cccc} 
\subfloat[$\eta = 0.1$]{\includegraphics[width=0.18\linewidth]{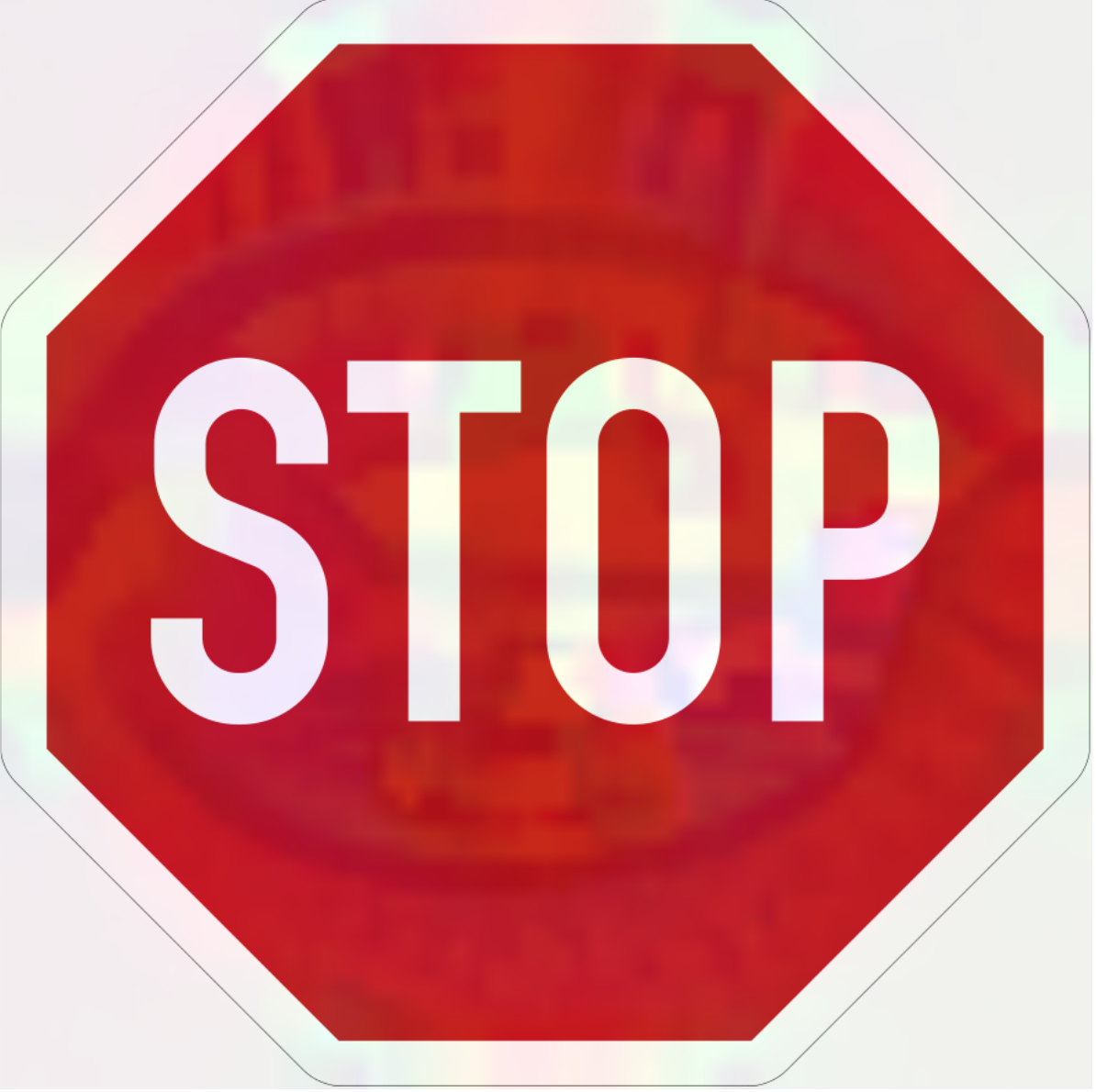}} \hspace{0.3em} &  
\subfloat[$\eta = 0.2$]{\includegraphics[width=0.18\linewidth]{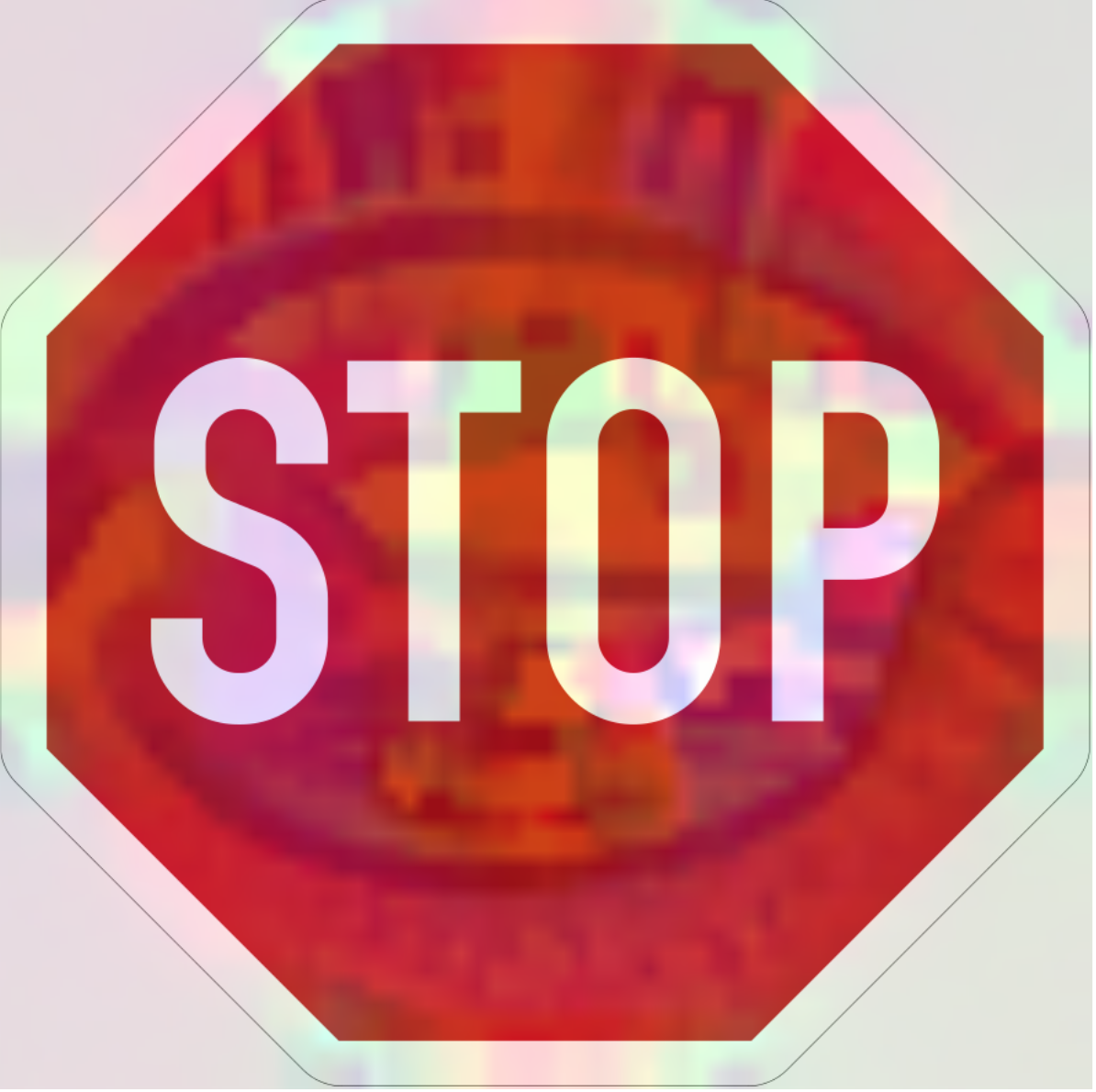}} \hspace{0.3em} &
\subfloat[$\eta = 0.3$]{\includegraphics[width=0.18\linewidth]{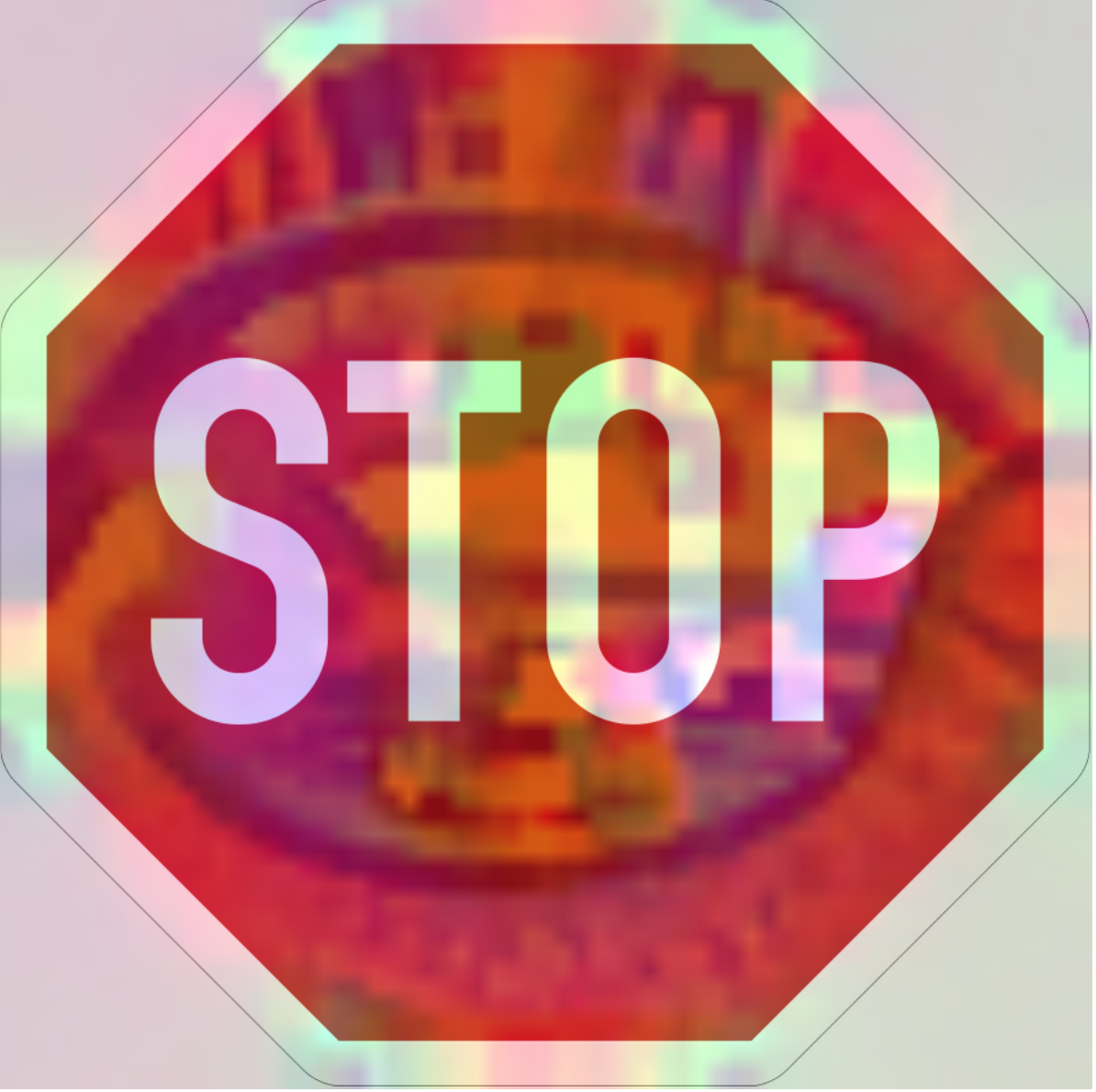}} \hspace{0.4em} &
\subfloat[$\eta=1$]{\includegraphics[width=0.18\linewidth]{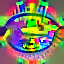}} \\
    \end{tabular}
{\includegraphics[width=0.98\linewidth]{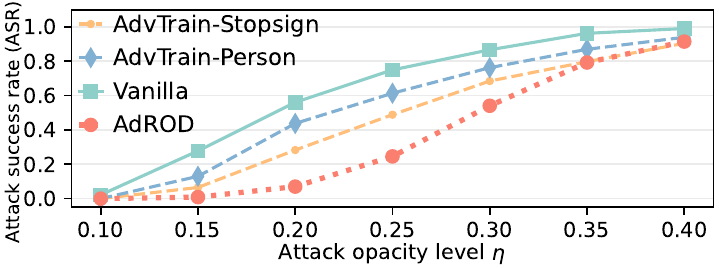}}
\vspace{-0.5em}
    \caption{ASR of global masking attacks targeting stop signs. Evaluated on the stop sign images from the outdoor testbed.}
    \label{fig:proj}
   \vspace{-1em}
\end{figure}

\textbf{Generalization.}  To further examine AdROD's generalizability beyond regional patches, we extend our evaluation to semi-transparent global masking attacks targeting stop signs as shown in Fig.~\ref{fig:proj}. We use two adversarial training configurations: \emph{AdvTrain-StopSign}, which is based on the correct target class but an incorrect masking type (i.e., solid regional masking during training but semi-transparent global masking during testing), and \emph{AdvTrain-Person}, which is based on an incorrect target class and solid regional masking type. Let $\eta$ denote the attack opacity. When $\eta \leq 0.3$, AdROD outperforms all baselines, including \emph{AdvTrain-StopSign}, which is otherwise highly effective against solid patch attacks. Beyond this threshold, high image distortion impairs all detectors. Among the two adversarial training baselines, \emph{AdvTrain-Person} performs noticeably worse, due to mismatches in both class and patch type. This reinforces the understanding of the limitations of adversarial training. That is, although it is effective when testing conditions match its training assumptions, its robustness declines when attack characteristics shift. In contrast, AdROD maintains robust performance across these masking variations, underscoring its better generalizability.

\subsubsection{Architectural Generalizability}

\label{sec:frcnn}

We extend AdROD to Faster R-CNN, a two-stage detector comprising a backbone with Feature Pyramid Networks (FPN), a Region Proposal Network (RPN), and a Region of Interest (RoI) head. We configure the low-rank HyperNetworks to update the weights of all convolutional layers in the backbone, FPN, and RPN, as well as the linear layers in the RoI head. This configuration yields a good balance between robustness and accuracy.

Unlike YOLO's single-stage architecture, Faster R-CNN filters low-score proposals at the RPN stage before classification. Consequently, the number and spatial distribution of final detections may not align across diverse ensemble members. To compute uncertainty for AdROD-I, we align the outputs by extracting objectness scores directly from the intermediate RPN logits and inserting low-score dummy bounding boxes wherever an anchor lacks a corresponding detection in any ensemble member. AdROD-II, however, operates purely on post-NMS temporal tracking and can be applied seamlessly.

 Table~\ref{tab:frcnn-results} shows that AdROD-I reduces ASR to about 0.09 across all settings with only a modest mAP drop, leaving little observable trend with respect to $K$. In contrast, AdROD-II becomes more robust as the serving ensemble size increases. At $K\!=\!4$, AdROD-II achieves the lowest ASR and surpasses AdROD-I. We attribute this improvement to AdROD-II's temporal dynamics-based attack detection, which enables more precise localization of adversarial patches and recoveries than AdROD-I's uncertainty-based estimation. Overall, these results confirm AdROD's generalizability to Faster R-CNN.

\begin{table}[t]
\centering
\caption{Performance (ASR and mAP50) of Faster R-CNN and AdROD defense variants under outdoor SysAdv attack.}
\label{tab:frcnn-results}
\setlength{\tabcolsep}{1pt}
\resizebox{0.98\linewidth}{!}{%
\renewcommand{\arraystretch}{1.0}
\begin{tabular}{c c ccc ccc}
\toprule
\multirow{2.5}{*}{\textbf{Meric}} & \textbf{No def.} & \multicolumn{3}{c}{\textbf{AdROD-I}} & \multicolumn{3}{c}{\textbf{AdROD-II}} \\ 
\cmidrule(lr){2-2} \cmidrule(lr){3-5} \cmidrule(lr){6-8}
& vanilla & $K=2$ & $K=3$ & $K=4$ & $K=2$ & $K=3$ & $K=4$ \\
\midrule
ASR & $0.635$ & $0.090$ & $0.098$ & $0.090$ & $0.137$ & $0.122$ & $0.071$ \\
mAP50 & $0.252$ & $0.223$ & $0.222$ & $0.222$ & \multicolumn{3}{c}{--} \\
\bottomrule
\end{tabular}%
}
\vspace{-1em}
\end{table}


\subsubsection{Adaptive Adversaries} 
\label{sec:Adaptive Adversaries}

We evaluate two adaptive attacks under a unified optimization protocol. The first, \emph{component-aware adaptive attack}, uses the standard hiding objective $\mathcal{L}_{\mathrm{hide}}$ (see \textsection\ref{detection_loss}), averaged over the vanilla detector and a surrogate ensemble constructed from exposed AdROD components. This averaging acts as a Monte Carlo approximation to an expectation over generated models and input transformations, while preserving the hiding effect on the vanilla detector. The second, \emph{uncertainty-aware adaptive attack}, further targets AdROD-I's uncertainty-aware selection rule by adding a disagreement loss that suppresses objectness variance across the surrogate ensemble. For both attacks, optimization is performed on differentiable detector outputs before NMS. At each step, the adversary constructs the surrogate ensemble according to its knowledge setting and applies the corresponding input transformations to the patched image.  After optimization, the patch is evaluated on the full defended pipeline. The adversary knows the defense pipeline and sampling distribution, but not the runtime noise vectors used by AdROD.


\textbf{Component-aware adaptive attack.}
Table~\ref{tab:Adaptive_Attack} reports the results under increasing attacker knowledge. The cases range from access only to the base detector $\Theta_0$ (\textbf{Case 0}, same as the setting in Table~\ref{tab:compare_digital}), to additional access to historical generated weights $\Theta$ (\textbf{Case 1}), to access to the HyperNetwork generator $\mathcal{G}$ for sampling surrogate runtime detectors (\textbf{Case 2}), and finally to full disclosure of all components (\textbf{Case 3}). We denote the surrogate ensemble size by $S$ and consider two settings: $S\!=\!2$, the smallest setting that still forms an ensemble, and $S\!=\!10$, matching the serving ensemble size of AdROD-I ($K=10$). We draw two key observations from Table~\ref{tab:Adaptive_Attack}. 
First, increasing the surrogate ensemble size reduces attack effectiveness.
Intuitively, a larger surrogate ensemble makes the optimization harder because the attacker must find one patch that hides the victim object across more generated models. 
Second, the optimized patches remain effective against the vanilla detector, especially when $S\!=\!2$. This shows that the adaptive attack can still produce valid hiding patches for the base model. However, the same patches transfer much less effectively to the full AdROD pipeline, indicating that AdROD's generated ensemble reduces attack transferability.
Even in the worst case observed in this table (\textbf{Case 2}, $S\!=\!2$), the ASR only increases moderately compared with the non-adaptive setting. These results show that AdROD remains robust against component-aware adaptive adversaries.

\begin{table}[t]
\centering
\caption{ASR of component-aware adaptive attack.}
\label{tab:Adaptive_Attack}
\begin{threeparttable}
\resizebox{0.98\linewidth}{!}{
\renewcommand{\arraystretch}{1.0}
\begin{tabular}{c *{3}{>{\centering\arraybackslash}p{1.9em}} cc cc}
\toprule
\multirow{2}{*}{\textbf{Case}} 
& \multicolumn{3}{c}{\textbf{Attacker Knowledge}} 
& \multicolumn{2}{c}{\textbf{Vanilla ASR}} 
& \multicolumn{2}{c}{\textbf{AdROD ASR}} \\ 
\cmidrule(lr){2-4} \cmidrule(lr){5-6} \cmidrule(lr){7-8}
& $\Theta_0$ & $\Theta$ & $\mathcal{G}$ 
& \multicolumn{1}{c}{\textbf{$S=2$}} 
& \multicolumn{1}{c}{\textbf{$S=10$}} 
& \multicolumn{1}{c}{\textbf{$S=2$}} 
& \multicolumn{1}{c}{\textbf{$S=10$}} \\ 
\midrule
\textbf{3} & \checkmark & \checkmark & \checkmark & $0.628$ & $0.234$ & $0.245$ & $0.169$ \\ 
\textbf{2} & \checkmark &            & \checkmark & $0.621$ & $0.222$ & $0.257$ & $0.153$ \\ 
\textbf{1} & \checkmark & \checkmark &            & $0.625$ & $0.326$ & $0.241$ & $0.184$ \\ 
\midrule
\textbf{0} & \checkmark &            &            & \multicolumn{2}{c}{$0.728$} & \multicolumn{2}{c}{$0.195$} \\ 
\bottomrule
\end{tabular}
}
\begin{tablenotes}
\footnotesize
\item[] $\Theta_0$: base detector; $\Theta$: a set of historically generated weights; 
\item[] $\mathcal{G}$: HyperNetwork generator and the random noise sampling distribution.
\item[] \parbox[t]{0.95\linewidth}{$S$ denotes the surrogate ensemble size, i.e., the number of generated models used by the adversary for patch optimization.}
\end{tablenotes}
\end{threeparttable}
\vspace{-1em}
\end{table}

\begin{table}[t]
\centering
\caption{ASR under uncertainty-aware adaptive attack.}
\label{tab:uncertainty_adaptive}
\begin{threeparttable}
\setlength{\tabcolsep}{3pt}
\resizebox{0.99\linewidth}{!}{
\renewcommand{\arraystretch}{1.}
\begin{tabular}{c ccccccc}
\toprule
\textbf{$\alpha_{\mathrm{dis}}$} & $0$ & $0.1$ & $0.5$ & $1$ & $5$ & $10$ & $100$ \\
\midrule
\textbf{Vanilla ASR} & $0.621$ & $0.617$ & $0.590$ & $0.582$ & $0.602$ & $0.594$ & $0.245$ \\
\textbf{AdROD ASR}  & $0.257$ & $0.218$ & $0.215$ & $0.207$ & $0.222$ & $0.215$ & $0.165$ \\
\bottomrule
\end{tabular}
}
\begin{tablenotes}
\scriptsize
\item[] \parbox[t]{1.0\linewidth}{The attack uses \textbf{Case 2} with $S=2$. $\alpha_{\mathrm{dis}}\!=\!0$ corresponds to the component-aware adaptive attack.}
\end{tablenotes}
\end{threeparttable} 
\vspace{-1.5em}
\end{table}

\textbf{Uncertainty-aware adaptive attack.}
We further evaluate a stronger white-box adversary that accounts for AdROD-I's uncertainty-aware selection rule. Specifically, we augment the original hiding loss $\mathcal{L}_{\mathrm{hide}}$ with a disagreement loss $\alpha_{\mathrm{dis}} \cdot \operatorname{Var}(\{o_i\}_{i=1}^{S})$, where $o_i$ denotes the victim object's objectness score from the $i$-th generated model in the surrogate ensemble, and $\alpha_{\mathrm{dis}}$ controls the weight of the disagreement loss. Since the attack minimizes this objective, the added term reduces objectness variance across the surrogate ensemble, making the attacked object less likely to be selected into $\mathcal{B}_{\mathrm{victim}}$ by AdROD-I. Based on Table~\ref{tab:Adaptive_Attack}, we use the strongest observed attack setting (\textbf{Case 2}, $S\!=\!2$), and sweep $\alpha_{\mathrm{dis}} \in \{0,0.1,0.5,1,5,10,100\}$. The case $\alpha_{\mathrm{dis}}\!=\!0$ corresponds to the component-aware adaptive attack under the same setting.

Table~\ref{tab:uncertainty_adaptive} shows that reducing objectness variance of the attacked object across the ensemble does not make the attack stronger against AdROD-I. Across all nonzero $\alpha_{\mathrm{dis}}$ values, the ASRs on AdROD are lower than in the $\alpha_{\mathrm{dis}}\!=\!0$ case. The vanilla ASR also drops when the disagreement loss becomes large, suggesting that overemphasizing disagreement reduction weakens the patch's basic hiding ability. The conflicting attack objectives, i.e., hiding the victim object and making the generated models respond similarly, impede optimization. These results further show that AdROD's diversity complicates adaptive patch optimization, even when the attacker accounts for AdROD-I's uncertainty-aware selection rule.

\subsection{An End-to-End Safety Evaluation}
\label{sec:e2e}

We employ OpenCDA~\cite{xu2021opencda}, a cooperative driving system integrated with the CARLA simulator~\cite{Dosovitskiy17}, to investigate the end-to-end effects of attack and defense in a specific scenario where the vehicle should stop before the stop line in front of a stop sign instrumented with adversarial patches. We implement a stop procedure as follows. When a stop sign is detected in at least three frames within a short time window and two of these frames are consecutive, the vehicle starts the stop procedure of decelerating to 5\,km/h and then engaging a longitudinal controller to halt before the stop line. The three-frame requirement handles transient FPs. There are three possible outcomes: (1) \emph{Safe stop}, where the vehicle halts entirely before the stop line; (2) \emph{Marginal stop}, where the vehicle halts on or slightly beyond the line; and (3) \emph{Missed stop}, where the vehicle fails to confirm the stop sign and bypasses braking. For safe and marginal stops, we record the remaining distance to the stop line as the {\em safety margin}. A comprehensive safety analysis covering all possible scenarios and adversarial attacks is beyond the scope of this paper. However, our results here provide insights into understanding the interplay among attack strength, defense, compute constraints, and safety.


The simulated vehicle initially cruises at a constant speed and executes the perception-control loop at a rate of 20\,Hz. Under no attack, with the vanilla model and the two AdROD variants, the vehicle achieves safe stops at all considered cruise speeds from 20 to 40\,km/h. This range allows us to evaluate stopping safety under increasingly demanding approach speeds, while remaining within a low-to-moderate speed regime relevant to urban road safety~\cite{campolettano2025potential}.



We use a SysAdv adversarial patch. We adjust only the lightness channel ($L$) of the patch in the CIELAB color space, while keeping the chrominance channels unchanged to preserve adversarial color features. The light-adjusted patch is derived by: $L_{\text{patch}}'=(1-\zeta)\cdot L_{\text{patch}}+\zeta\cdot L_{\text{matched}}$,  where $L_{\text{patch}}$ is the original lightness of the patch, and $L_{\text{matched}}$ is obtained by linearly scaling and shifting $L_{\text{patch}}$ to match the mean and standard deviation of the ambient lightness. Thus, $\zeta=0$ leaves the patch unchanged, while $\zeta=1$ fully matches the ambient lightness statistics. We use $\zeta=0.5$ by default.



\begin{figure}
    \centering
    \includegraphics[width=1\linewidth]{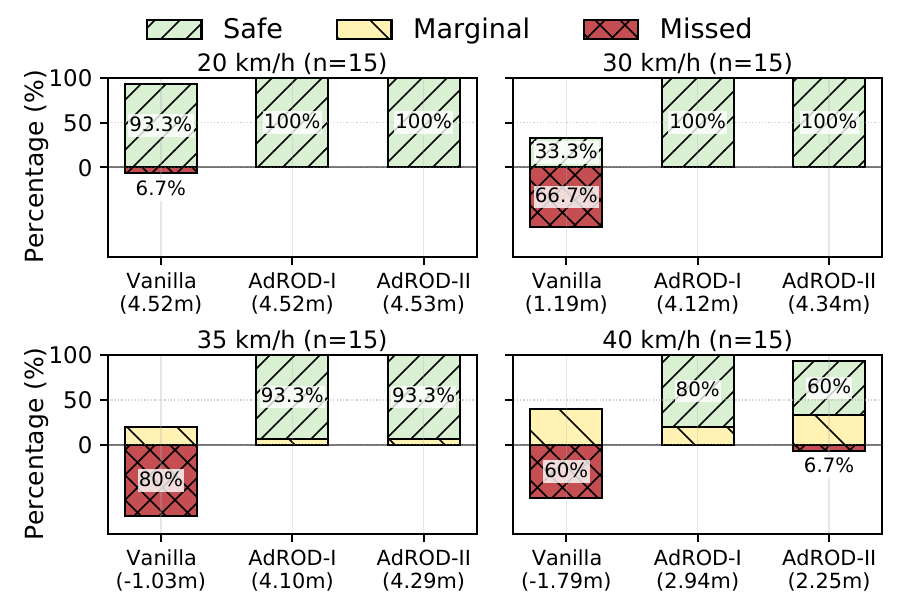}
    \caption{Stop outcomes in front of an adversarial stop sign. Bars show the percentages of \emph{safe}, \emph{marginal}, and \emph{missed} stops. Values below the x-axis give the mean signed distance to the stop line, where positive values indicate stopping before the line and negative values indicate stopping beyond it. $n$ denotes the number of trials per bar.}
    
    \label{fig:stop_e2e}
    \vspace{-1em}
\end{figure}

\subsubsection{Safety Under Attack}
Fig.~\ref{fig:stop_e2e} shows that the undefended vanilla model's performance rapidly degrades as the vehicle speed increases. In contrast, both AdROD variants yield no missed stops up to 35\,km/h. At 40\,km/h, AdROD-I still yields no misses, whereas AdROD-II experiences a rare missed stop. This occurs because the high vehicle speed limits the number of frames containing the stop sign, impeding AdROD-II in establishing the kinematic tracking of the attacked sign for defense activation. Note that if we set $\zeta > 0.5$, the increased lightness matching will reduce the attack strength; however, the vanilla model still yields substantial missed stop rates at 40\,km/h across all tested $\zeta$ values (e.g., over 50\% at $\zeta = 0.6$ and above 10\% even at $\zeta = 1.0$). In contrast, both AdROD variants achieve zero missed stops across these setups.



 Fig.~\ref{fig:stop_e2e} also shows that both AdROD variants leave substantially larger safety margins than the vanilla model. In addition, they excel at different speeds. At moderate speeds (20--35\,km/h), AdROD-II achieves slightly larger margins, as once the ensemble defense is triggered, it can precisely and continuously track the stop sign and invoke the stop procedure. As for AdROD-I, if the attack does not immediately induce a significant disagreement, stop sign recovery as well as the stop procedure may be delayed. However, at 40\,km/h, this trend reverses. The delayed activation of AdROD-II, which confirms an adversarial scenario after $\ell$ frames of loss of the tracked sign, consumes greater braking distance at high speeds. In contrast, AdROD-I operates continuously and produces earlier braking in these experiments.
 


\subsubsection{Robustness Under Compute Constraints}

\begin{figure}[t]
    \centering
    \includegraphics[width=1\linewidth]{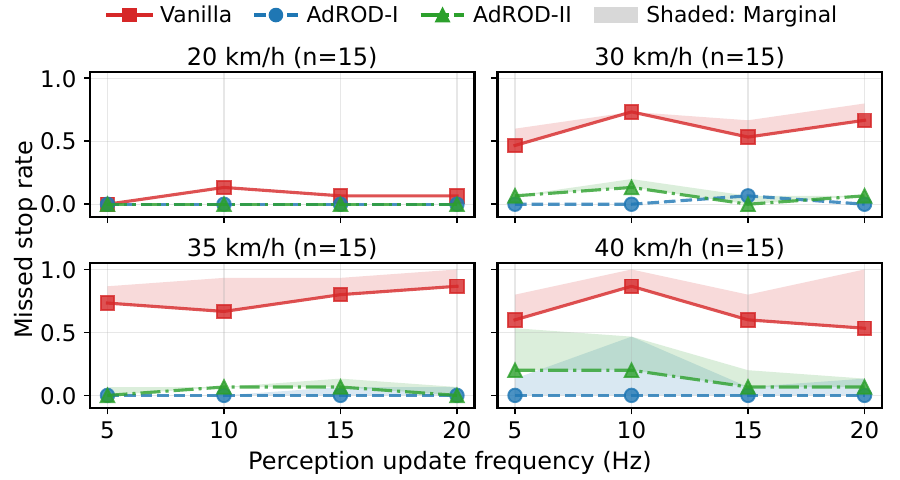}
    \caption{
    Impact of reduced perception update frequency due to computational constraints on stopping outcomes. Solid lines denote the rate of missed stops; shaded bands represent the rate of marginal stops. $n$ denotes the number of trials per data point.}
    \label{fig:compute_budget}
    \vspace{-1em}
\end{figure}

The earlier results are obtained at a perception-control rate of 20\,Hz. To understand how computational constraints, such as those caused by resource contention, affect the safety of the undefended and defended systems,
we simulate the scenario where the perception (including object detection) is executed at lower rates but the vehicle control is still performed at 20\,Hz.
To deal with the difference between their execution rates, the latest perception results are used by the vehicle control.

Fig.~\ref{fig:compute_budget} shows the rates of missed stop and marginal stop versus the perception rate under various vehicle cruise speeds. With the undefended vanilla model, the two rates are mainly affected by the vehicle speed, rather than the perception rate.

The AdROD variants effectively suppress missed stops. At 20--30\,km/h, missed and marginal stop rates remain near zero across all perception rates. Degradation becomes pronounced only at 40\,km/h, where AdROD-I maintains zero missed stops, while AdROD-II shows increased missed and marginal stops at 5--10\,Hz. This is because high vehicle speeds combined with reduced perception rates weaken the temporal continuity available for object tracking, making it more difficult for AdROD-II to detect an abrupt disappearance and activate the defense in time. Nevertheless, AdROD-II's near-baseline overhead of 1.03$\lambda$ leaves greater computational headroom for maintaining the perception rate, although the realized rate under resource contention remains platform-dependent.

\section{Discussion}
\label{discussion}

AdROD focuses on hiding attacks against genuine objects in camera-based object detection. It does not address attacks that fabricate \textit{fake objects}, such as phantom signs~\cite{nassi2020phantom}, which may even deceive human observers. Defending against object-creation attacks would require additional mechanisms, such as object existence verification or contextual reasoning. AdROD is also specialized for bounding-box object detectors. While the idea of stochastic model generation with functional diversity could be extended to other perception tasks, such as semantic segmentation or instance segmentation, doing so would require task-specific output fusion, uncertainty estimation for AdROD-I, and activation logic for AdROD-II. We leave these extensions as future work. Finally, AdROD-II may fail to activate if an attack-instrumented object first appears when the patch is already effective and remains undetected thereafter, leaving no prior track to trigger the ensemble. Although maintaining such continuous suppression in practice is difficult, we identify this as a limitation of AdROD-II. Periodic ensemble checks can serve as a practical fallback.

\section{Conclusion}  
\label{dis_and_conclude}  

This paper presented AdROD, a stochastic ensemble defense for camera-based object detection in autonomous driving. It includes an efficient HyperNetwork training framework and introduces functional diversity to enhance defense performance. With two serving modes, AdROD supports different deployment goals: robust detection in AdROD-I and runtime-efficient defense in AdROD-II. Across diverse attack settings, AdROD outperforms representative baseline defenses and generalizes better than evaluated adversarial-training baselines across the tested shifts.





\section*{Acknowledgment}
The authors used OpenAI ChatGPT solely for language editing and polishing of author-written text. It was not used to generate scientific ideas, methods, results, figures, or code. All revisions were reviewed and verified by the authors.

{\footnotesize
\bibliographystyle{IEEEtran}
\bibliography{sample-base}
}

\begin{IEEEbiography}[{\includegraphics[width=1in,height=1.25in,clip, keepaspectratio]{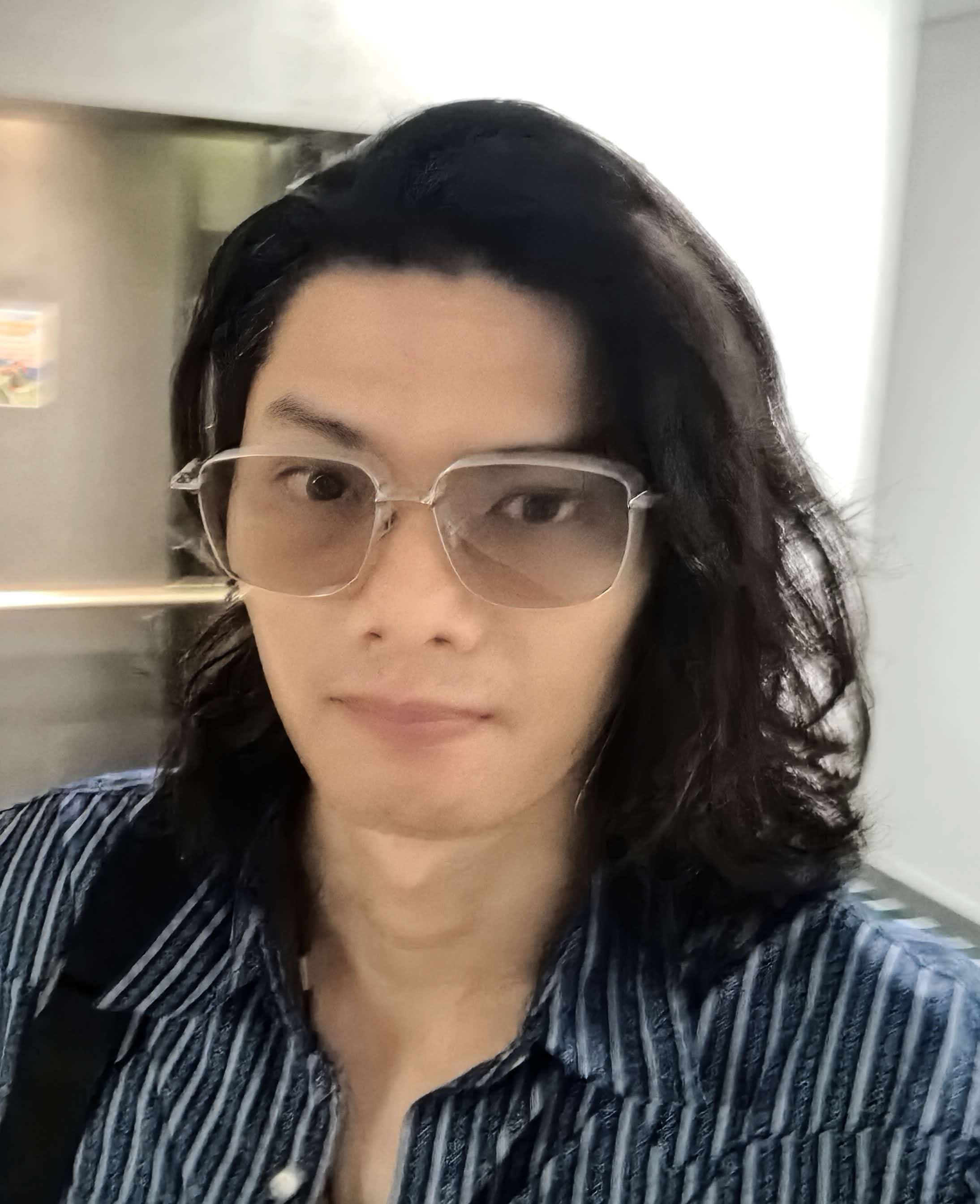}}]{Yuting Wu}
received the B.Eng. degree from Northeastern University, China, in 2019, and the M.Sc. degree from Nanyang Technological University (NTU), Singapore, in 2020. From 2020 to 2026, he was a Research Associate at NTU. He is currently pursuing the Ph.D. degree with the College of Computing and Data Science, NTU. His research interests include autonomous driving, reinforcement learning, adversarial attacks, and world models.
\end{IEEEbiography}

\vspace{-1.5\baselineskip}

\begin{IEEEbiography}[{\includegraphics[width=1in,height=1.25in,clip,keepaspectratio]{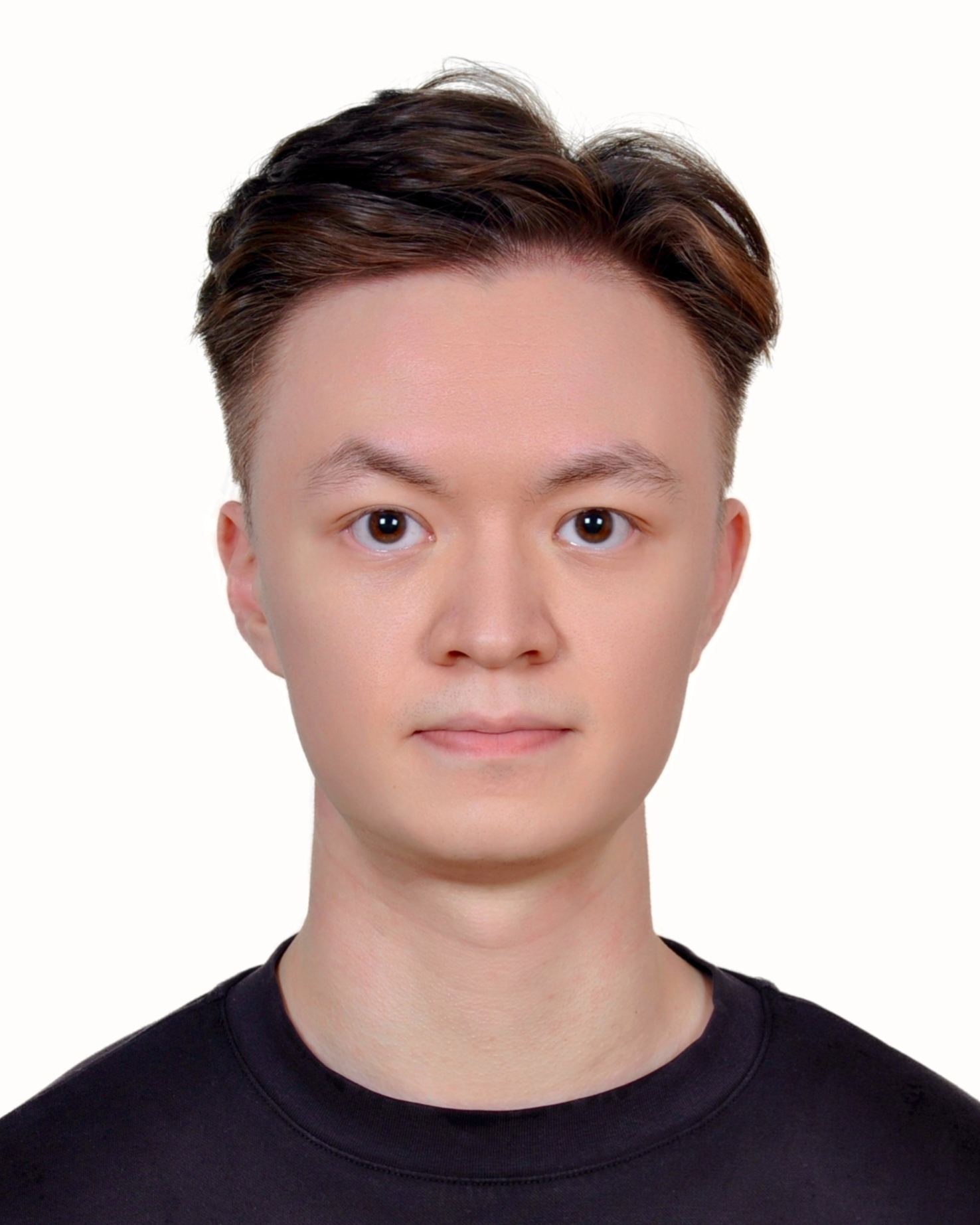}}]{Dongfang Guo}
received the B.Eng. degree from the University of Electronic Science and Technology of China, China, and the M.Sc. and Ph.D. degrees from Nanyang Technological University, Singapore. He is currently a Research Fellow with the College of Computing and Data Science, NTU. His research interests include reliable and secure sensing for the Internet of Things and cyber-physical systems, with particular interests in robust perception for autonomous systems and resource-aware edge AI.
\end{IEEEbiography}

\vspace{-1.5\baselineskip}

\begin{IEEEbiography}[{\includegraphics[width=1in,height=1.25in,clip,keepaspectratio]{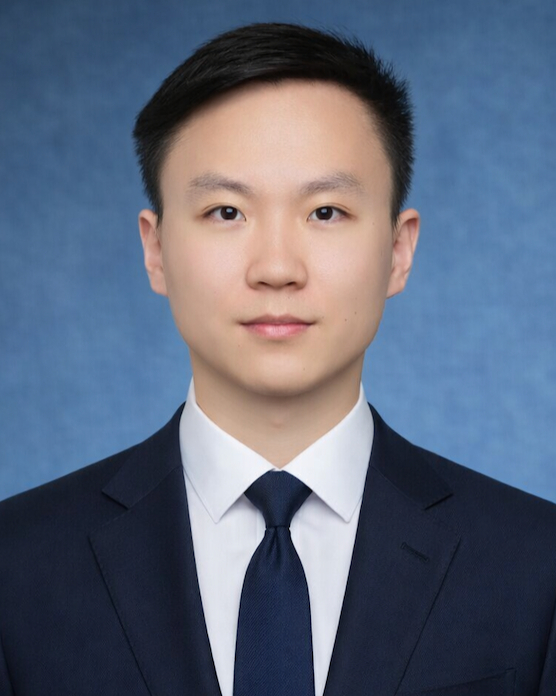}}]{Xiangzhong Luo}
received the B.E. degree from Shanghai Jiao Tong University, China, in 2019, and the Ph.D. degree from Nanyang Technological University, Singapore, in 2023. From 2024 to 2025, he was a Research Fellow at Nanyang Technological University. He is now an Associate Professor at the School of Computer Science and Engineering, Southeast University, China. His research interests include embedded systems, neural architecture search, model compression, and LLM inference acceleration.
\end{IEEEbiography}

\vspace{-1.5\baselineskip}

\begin{IEEEbiography}[{\includegraphics[width=1in,height=1.25in,clip,keepaspectratio]{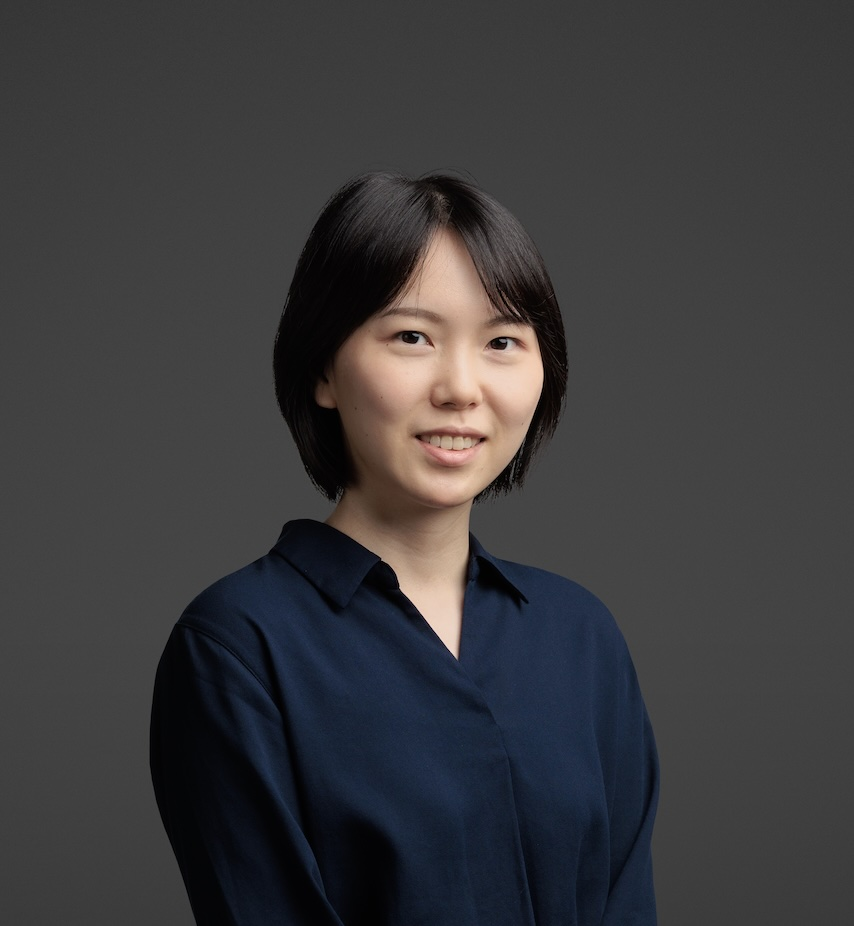}}]{Qun Song}
received the B.Eng. degree from Nankai University, China, and the Ph.D. degree from Nanyang Technological University, Singapore. She is currently an Assistant Professor with the Department of Electrical Engineering, City University of Hong Kong. From 2022 to 2024, she was an Assistant Professor with the Embedded Systems Group, Faculty of Electrical Engineering, Mathematics and Computer Science, Delft University of Technology, the Netherlands, and from 2024 to 2025, with the Information Systems Technology and Design Pillar, Singapore University of Technology and Design. Her research interests include Artificial Intelligence of Things (AIoT), cyber-physical system robustness and resilience, and autonomous driving security and safety. Her honors include the 2023 MobiCom Best Community Contribution Award, the 2022 SenSys Best Paper Award Finalist, and the 2021 IPSN Best Artifact Award Runner-up.
\end{IEEEbiography}

\vspace{-1.5\baselineskip}

\begin{IEEEbiography}[{\includegraphics[width=1in,height=1.25in,clip,keepaspectratio]{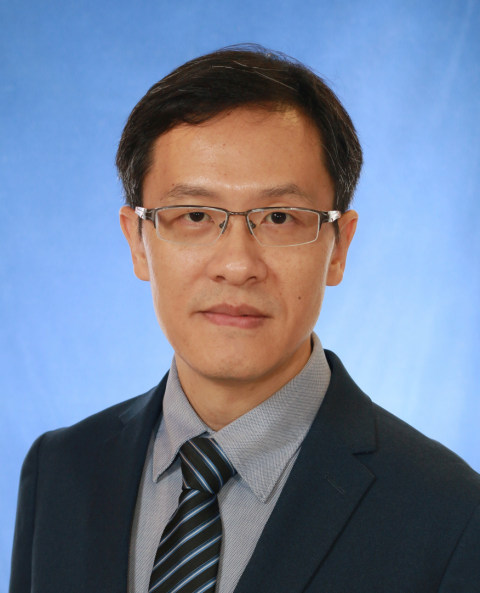}}]{Rui Tan} is a Full Professor at College of Computing and Data Science, Nanyang Technological University, Singapore. Previously, he was a Research Scientist (2012-2015) and a Senior Research Scientist (2015) at Advanced Digital Sciences Center of University of Illinois at Urbana-Champaign, and a postdoctoral Research Associate (2010-2012) at Michigan State University. He received the Ph.D. (2010) degree in computer science from City University of Hong Kong, the B.S. (2004) and M.S. (2007) degrees from Shanghai Jiao Tong University. His research interests include cyber-physical systems and Internet of things. He is the recipient of Best Demo Award from SenSys’24, Best Paper Awards from ICCPS’23 and IPSN’17. He served as Associate Editor of IEEE Transactions on Mobile Computing and ACM Transactions on Sensor Networks, TPC Co-Chair of e-Energy’23, EWSN’24, SenSys’24, and General Co-Chair of e-Energy’24 and RTCSA’25. He received the Distinguished TPC Member recognition from SenSys in 2025 and from INFOCOM in 2017, 2020, and 2022.
\end{IEEEbiography}

\end{document}